\documentclass[aip, jcp, preprint]{revtex4-1}

\usepackage{color}
\usepackage{amsmath}
\usepackage{graphicx}
\usepackage{amssymb}
\usepackage{comment}
\usepackage{url}
\usepackage{algorithm}

\usepackage[normalem]{ulem}     

\begin{document}

\title{Sparse learning of stochastic dynamic equations}

\date{\today}

\author{Lorenzo Boninsegna}
\affiliation{Department of Chemistry and Center for Theoretical Biological Physics, Rice University}

\author{Feliks N\"{u}ske}
\affiliation{Department of Chemistry and Center for Theoretical Biological Physics, Rice University}

\author{Cecilia Clementi}
\affiliation{Department of Chemistry and Center for Theoretical Biological Physics, Rice University}
\email{cecilia@rice.edu}

\date{\today}

\begin{abstract}
With the rapid increase of available data for complex systems, there is great
interest in the extraction of physically relevant information from massive
datasets.
Recently, a framework called Sparse Identification of Nonlinear Dynamics
(SINDy) has been introduced to identify the governing equations of dynamical
systems from simulation data. In this study, we extend SINDy to stochastic
dynamical systems, which are frequently used to model biophysical processes. We
prove the asymptotic correctness of stochastics SINDy in the infinite data
limit, both in the original and projected variables. We discuss algorithms to
solve the sparse regression problem arising from the practical implementation
of SINDy, and show that cross validation is an essential tool to determine the
right level of sparsity. We demonstrate the proposed methodology on two test
systems, namely, the diffusion in a one-dimensional potential, and the
projected dynamics of a two-dimensional diffusion process.
\end{abstract}


\maketitle 




\section{Introduction}
The last decade has seen a dramatic increase in our ability to collect or
produce large amounts of high resolution and high dimensional data associated
with complex physical and chemical systems, both by means of experimental
measurements or computer simulations. In many different scientific fields,
ranging from high energy physics to neuroscience, the ``big-data'' problem has
spurred interest in data analysis methods that can condense massive datasets
into a minimal amount of essential information and/or can detect
relevant patterns and anomalies in the distribution of the data.

In the specific case of molecular systems, a large body of work has been
devoted to define collective coordinates and reaction pathways from molecular
dynamics simulation data
\cite{NoeClementiCOSB2017,RohrdanzEtAl_AnnRevPhysChem13_MountainPasses}. 
However, most of the proposed techniques are descriptive and do
not provide a functional link relating the variables to the observed behavior.
Mathematical approaches that have been proved optimal to reduce the complexity
of the data by dimensionality reduction and/or coarse graining (in time or
space) usually do not offer a straightforward physical interpretation of the
results. Here we take a different approach and make a first step towards the
definition of methods to learn the functional form of a molecular model from the
available data. 

Assuming an extensive sampling of a given set of variables describing a system is
available for a certain time frame, different data-driven methods have been
proposed to ``learn'' how to propagate the system to future times, either in
terms of the original variables or in a reduced representation.
For instance, the so-called ``equation-free'' approach, uses local (in time and
space) microscopic simulations to propagate macroscopic variables to long
timescales~\cite{Kevrekidis2009}. Such an approach bypasses the need of formulating
constitutive equations for the time evolution of the macroscopic variables of the
system in closed form and provides a practical recipe for multiscale simulation.
However, it is oftentimes desirable to obtain an explicit analytical expression
for the dynamical equations in terms of the variables of interest, as they can
offer a physicochemical understanding of the system.
Ideally, one would like to design approaches that are able to infer such
equations from the available data.
Recently, a significant step in this direction has been proposed for
deterministic dynamical systems~\cite{BruntonPNAS2016}. The Sparse
Identification of Nonlinear Dynamics (SINDy) approach combines ideas from sparse
regression~\cite{Tibshirani96Lasso, James2013} and compressed
sensing~\cite{Donoho2006, Baraniuk2007} to automatically discover the terms of the
differential equations (either ordinary~\cite{BruntonPNAS2016} or
partial~\cite{Rudye1602614}) that best represent large sets of time-dependent data, 
given a suitable function library (as it will be discussed below).  For
instance, it was shown that SINDy can be used to obtain the correct equation
for the low-dimensional slow attractor associated with the dynamics of a fluid
flow past a cylinder, that is described by the Navier-Stokes equations at the
microscopic scale~\cite{BruntonPNAS2016}.

Such a methodology appears very promising to learn effective equations
of motion in different fields of application, such as molecular systems. A
significant difference that limits the application of SINDy to (macro)molecular
systems is the presence of noise, as their dynamics are usually non-deterministic.
Towards this goal, here we present an extension of this approach that allows to
derive stochastic dynamical equations from data, either to describe the time
evolution of microscopic variables or of their transformation in a different
space. For the latter, we combine the SINDy idea with the formalism of
projected stochastic dynamics \cite{Legoll_Nonlinearity2010, Zhang_Faraday16}. 
We show that extensive cross-validation is a crucial ingredient that needs to
be added in the sparsification of the solution for this approach to be
successful in the presence of significant noise and/or limited data. 

The manuscript is organized as follows. First, the proposed extension of SINDy
to stochastic systems and its theoretical underpinnings are outlined. We show
how trajectory data can be used as an input to formulate a regression problem
approximating the drift and diffusion coefficients of an Ito process, both in
the microscopic and in an effective variable space. The specific algorithm used
to solve the regression is then detailed, by introducing a cross validation
based Stepwise Sparse Regression. Such a formalism is employed to learn
dynamical equations from data for two test systems: the homogeneous diffusion in
a one dimensional double well potential, and the projected dynamics along a
projected coordinate in a two dimensional potential. Results and implications
are finally discussed.

\section{Theory}

\subsection{Sparse Identification of Dynamical Systems}
\label{sec:SINDy}
We start by outlining the SINDy approach for deterministic dynamical
systems, that was originally proposed in ref.~\citenum{BruntonPNAS2016}. The goal is to
learn the dynamical equations for a system described by an ordinary
differential equation:

\begin{eqnarray}
\frac{d}{dt}X(t) & = & F(X(t)),\label{eq:dynamical_system}
\end{eqnarray}
where $X(t)\in\mathbb{R}^{d}$ is the state of the system at time
$t$ and $F:\,\mathbb{R}^{d}\rightarrow\mathbb{R}^{d}$ is the vector
field defining the dynamics. For many complex systems, no closed-form
expression for the vector field $F$ is known, and the process can only
be observed through simulation or measurement data $X(t_{l}),\,l=1,\ldots,N$,
where $t_{1}<\ldots<t_{N}$ are discrete points in time. However,
it was suggested in ref.~\citenum{BruntonPNAS2016} to learn the dynamical
equation as a linear combination of a pre-selected dictionary of basis
functions. More precisely, let $\Theta_K = (f_1, \ldots, f_K)$ be a set of $K$
user-defined trial functions. Making the \emph{ansatz}:
\begin{eqnarray}
F_{i} & = & \sum_{k=1}^{K}c{}_{i,k}f_{k},
\label{eq:ansatz_vector_fields}
\end{eqnarray}
for the $i$-th component of the vector field, one arrives at a system
of $N$ linear equations for each time step by inserting Eq.~(\ref{eq:ansatz_vector_fields})
into Eq.~(\ref{eq:dynamical_system}):
\begin{eqnarray}
\frac{d}{dt}X_{i}(t_{l}) & = & \sum_{k=1}^{K}c_{i,k}f_{k}(X(t_{l})).
\label{eq:derivatives_lc_basis}
\end{eqnarray}
If the time derivatives on the left hand side of Eq.~(\ref{eq:derivatives_lc_basis})
can be computed, this defines a linear system

\begin{eqnarray}
\mathbb{Y}_i & = & \mathbb{X} \mathbf{c}_i,
\label{eq:linear_system_dyn_system}
\end{eqnarray}
where $\mathbb{Y}_i \in\mathbb{R}^{N}$ contains the  time derivatives at all
sampled time steps,
$\mathbb{X} \in\mathbb{R}^{N\times K}$ contains the evaluations of all
basis functions in dictionary $\Theta_K$ at all time steps, and $\mathbf{c}_i \in\mathbb{R}^{K}$ is
the unknown vector of coefficients:

\begin{eqnarray*}
Y_{i,l} & = & \frac{d}{dt}X_{i}(t_{l}),\\
X_{l,k} & = & f_{k}(X(t_{l})).
\end{eqnarray*}
Eq.~(\ref{eq:linear_system_dyn_system}) needs to be solved in
the least-squares sense, that is, $\tilde{\mathbf{c}}_i$ becomes the minimizer of

\begin{eqnarray}
\tilde{\mathbf{c}}_i & = & \min_{\mathbf{c}_i \in\mathbb{R}^{K}}  \|\mathbb{Y}_i-\mathbb{X}  \cdot \mathbf{c}_i \|_{2}^{2}.
\label{eq:least_squares_dyn_system}
\end{eqnarray}
In general, the solution $\tilde{\mathbf{c}}_i$ of Eq.~(\ref{eq:least_squares_dyn_system})
will not be sparse. 
If the goal is to find the functional form of the vector field that better
represents the data among the large number of possibilities offered in the
function dictionary, sparsity of  $\tilde{\mathbf{c}}_i$ needs to be enforced.
Formally, this can be achieved by penalizing the $L^{1}$-norm of the solution
and minimizing
\begin{eqnarray}
\tilde{\mathbf{c}}_i & = & \min_{\mathbf{c}_i \in\mathbb{R}^{K}} ( \|\mathbb{Y}_i-\mathbb{X} \mathbf{c}_i \|_{2}^{2}+\rho\|\mathbf{c}_i\|_{1} ),
\label{eq:least_squares_dyn_system_sparse}
\end{eqnarray}
using some positive Lagrange multiplier $\rho$ which controls the weight of
the sparsity constraint. Algorithms to solve Eq.~(\ref{eq:least_squares_dyn_system_sparse})
will be discussed below.

\subsection{Sparse Identification of Stochastic Dynamics}

\paragraph{Diffusion Processes}

In this work, we extend the sparse learning framework discussed above
to stochastic dynamics. Instead of Eq.~(\ref{eq:dynamical_system}),
we consider dynamics driven by an Ito stochastic differential equation
(SDE)
\begin{eqnarray}
dX(t) & = & b(X(t))dt+\sqrt{2\beta^{-1}}\sigma(X(t))dW_{t}.
\label{eq:ito_sde}
\end{eqnarray}
Again, $X(t)\in\mathbb{R}^{d}$ denotes the state of the system at
time $t$, while $b:\,\mathbb{R}^{d}\mapsto\mathbb{R}^{d}$ is a vector
field called the \emph{drift}, and $\sigma:\,\mathbb{R}^{d}\mapsto\mathbb{R}^{d\times d}$
is a matrix field called the \emph{diffusion} of the process. Moreover, $W_{t}$
denotes a $d$-dimensional Brownian motion, and the prefactor $\beta$
is the inverse temperature $\beta = \frac{1}{k_B T}$ in physical applications.
The covariance
matrix of the diffusion is commonly denoted by $a\in\mathbb{R}^{d\times d}$,
\begin{eqnarray*}
a(x) & = & \sigma^{T}(x)\sigma(x).
\end{eqnarray*}
We will also refer to a process like Eq.~(\ref{eq:ito_sde}) as a
diffusion process. We assume the process $X(t)$ to be ergodic w.r.t.
a unique invariant measure $\mu$. 

A familiar variant of such a process is the overdamped Langevin dynamics
\begin{eqnarray}
dX(t) = -\frac{\nabla U(X(t))}{\gamma}dt+\sqrt{2\beta^{-1} \gamma}\, dW_{t},
\label{eq:overdamped_langevin}
\end{eqnarray}
that is, the drift is the gradient of a potential energy
function $U$ (the force) normalized by the friction coefficient
$\gamma$, while the diffusion matrix is constant. The equilibrium distribution
associated to this dynamical process is the Boltzmann distribution: $\mu(x) \propto \exp{(-\beta U(x))}$.

A diffusion process is thus generally defined by two components, the drift and
the diffusion. Both of them can be estimated from data via the Kramers-Moyal
expansion~\cite{Risken1989}:

\begin{eqnarray}
b_{i}(x) & = & \lim_{s\rightarrow0}\mathbb{E}\left[\frac{1}{s}(X_{i}(s)-x_{i})\vert X(0)=x\right],\label{eq:Kramers_Moyal_1}\\
a_{ij}(x) & = & \frac{\beta}{2}\lim_{s\rightarrow0}\mathbb{E}\left[\frac{1}{s}(X_{i}(s)-x_{i})(X_{j}(s)-x_{j})\vert X(0)=x\right].
\label{eq:Kramers_Moyal_2}
\end{eqnarray}
The expectations above average the linear and quadratic variation
of the process $X(s)$, conditioned on starting at position $x$ at
time $s=0$. 

If the linear and quadratic time variations on the right hand sides of
Eq.~(\ref{eq:Kramers_Moyal_1}) and Eq.~(\ref{eq:Kramers_Moyal_2}) can be computed, a
regression problem analogous to Eq.~(\ref{eq:least_squares_dyn_system}) can be
formulated and both drift and diffusion can be approximated as an optimal
linear combinations of basis functions.

\paragraph{Projected Dynamics}

In many physical applications, a diffusion process is not observed through its
original state space (e.g. atomic coordinates), but
through a projected space of lower dimension (e.g. dihedral angles or
interatomic distances in macromolecular dynamics). In this case,
it is desirable to learn a stochastic dynamical system defined only
along the projected variables from the data, often called an effective
dynamics, while discarding the other features. We now investigate this issue by
following the projection formalism previously proposed in other works~\cite{Legoll_Nonlinearity2010, Zhang_Faraday16}. 
It is important to note that
there are many possible ways of defining an effective dynamics on projected
variables (see ref.~\citenum{Zhang_Faraday16} for a discussion). In
practice, an effective dynamics in the form of an Ito stochastic differential
equation (that is, without memory terms) is meaningful if the projected
variables capture the slowest dynamical processes and a separation of
timescales exists in the system. In the following we assume this to be the case. 

Assume the projection is realized by a map $\xi:\,\mathbb{R}^{d}\mapsto\mathbb{R}^{m},\,m\leq d$,
and denote points in the projected space by $z\in\mathbb{R}^{m}$.
The level set of a point $z$ is denoted by

\begin{eqnarray*}
\Sigma_{z} & = & \left\{ x\in\mathbb{R}^{d}:\,\xi(x)=z\right\} .
\end{eqnarray*}
The projected stationary distribution is obtained by averaging the equilibrium
distribution $\mu$ over the level sets $\Sigma_{z}$:

\begin{eqnarray*}
\nu(z) & = & \int_{\Sigma_{z}}\mu(x)J^{-1/2}(x)\,\mathrm{d}\sigma_{z}(x),
\end{eqnarray*}
where $J$ is the Jacobian determinant of the transformation $\xi$,
and $\sigma_{z}$ denotes the surface measure on the manifold $\Sigma_{z}$.
It can be shown~\cite{Zhang_Faraday16} that $\nu$ defines a probability measure on the low-dimensional
space $\mathbb{R}^{m}$. Also, we can define a probability measure
$\mu_{z}$ which restricts the equilibrium measure to a level set
by

\begin{eqnarray}
\mathrm{d}\mu_{z}(x) & = & \frac{1}{\nu(z)}\mu(x)J^{-1/2}(x)\,\mathrm{d}\sigma_{z}(x)
\label{eq:restricted_eq_measure}
\end{eqnarray}
for $x\in\Sigma_{z}$. Like in the previous section, an effective
dynamics on the lower-dimensional space $\mathbb{R}^{m}$ can now
be defined by using the Kramers-Moyal expansion:
\begin{eqnarray}
b_{i}^{\xi}(z) & = & \lim_{s\rightarrow0}\mathbb{E}\left[\frac{1}{s}(\xi_{i}(X(s))-z_{i})\vert X(0)\sim\mu_{z}\right],\label{eq:Kramers_Moyal_eff_1}\\
a_{ij}^{\xi}(z) & = & \frac{\beta}{2}\lim_{s\rightarrow0}\mathbb{E}\left[\frac{1}{s}(\xi_{i}(X(s))-z_{i})(\xi_{j}(X(s))-z_{j})\vert X(0)\sim\mu_{z}\right].\label{eq:Kramers_Moyal_eff_2}
\end{eqnarray}
The difference between these and Eqs.~(\ref{eq:Kramers_Moyal_1}-\ref{eq:Kramers_Moyal_2})
is that the dynamics is observed along the projection $\xi$ here, and
that the initial condition is replaced by starting the process from
the distribution $\mu_{z}$ instead of starting deterministically
at one point.

\paragraph{Convergence Result}

Just as in section~\ref{sec:SINDy}, we would like to model the components
of the (effective) drift and diffusion terms by a linear combination of
pre-selected basis functions. We show that, given equilibrium
simulation data, we only need to compute the linear and quadratic
variations for all data points and approximate these data by a linear
regression, as the equilibrium sampling automatically takes care of the
averages required in Eqs.~(\ref{eq:Kramers_Moyal_eff_1}-\ref{eq:Kramers_Moyal_eff_2}).
This is the essence of the following convergence result, which we
prove in Appendix~\ref{sec:proof_thm1}. We introduce the following 

\textbf{Theorem 1:} \textit{Let $\left\{ X(t_1), \cdots, X(t_{N+1}) \right\}$ be a
$d$-dimensional time series from a diffusion process as Eq.~(\ref{eq:overdamped_langevin}) or
Eq.~(\ref{eq:ito_sde}), sampled with an uniform time window $s$. Furthermore, let 
$\Theta = \left( f_{1},\ldots,f_{K} \right)$ be a dictionary of basis functions on the projected space $\mathbb{R}^{m}$.
Define the database matrix $\mathbb{X} = \Theta(X(t_{l})) \in\mathbb{R}^{N\times K}$ and introduce
 the set of vectors $\mathbb{Y}_i,\,\mathbb{Y}_{ij}\in\mathbb{R}^{N}$ as:}
\begin{eqnarray}
Y_{i,l} & = & \frac{1}{s}\left[\xi_{i}(X(t_{l+1}))-\xi_{i}(X(t_{l}))\right]  \label{eq:simple_increm}\\
Y_{ij,l} & = & \frac{\beta}{2}\frac{1}{s}\left[\xi_{i}(X(t_{l+1}))-\xi_{i}(X(t_{l}))\right]\left[\xi_{j}(X(t_{l+1}))-\xi_{j}(X(t_{l}))\right]  \label{eq:squared_increm}
\end{eqnarray}
\textit{$\forall i, j = 1,\dots,m$. Then, as $N\rightarrow\infty,\,s\rightarrow0$, the solutions
$\left\{\tilde{\mathbf{c}}_{i}, \, \tilde{\mathbf{c}}_{ij} \right\} \in \mathbb{R}^K$ of the regression problems}
\begin{eqnarray}
\tilde{\mathbf{c}}_{i} & = & \min_{\mathbf{c}_i \in\mathbb{R}^{K}}\|\mathbb{Y}_{i}-\mathbb{X} \mathbf{c}_i\|_{2}^{2},\quad \forall i =1 , \cdots, m \label{eq:regression_drift}\\
\tilde{\mathbf{c}}_{ij} & = & \min_{\mathbf{c}_{ij} \in\mathbb{R}^{K}}\|\mathbb{Y}_{ij}-\mathbb{X} \mathbf{c}_{ij}\|_{2}^{2}, \quad \forall i,j =1 , \cdots, m\label{eq:regression_diffusion}
\end{eqnarray}
\textit{converge to the coefficient vectors of the best approximation
problems}
\begin{eqnarray*}
\tilde{\mathbf{c}}_{i} & = & \min_{\mathbf{c}_i \in\mathbb{R}^{K}}\|b^{\xi}_{i}-\sum_{k=1}^{K}c_{i, k}f_{k}\|_{L_{\nu}^{2}}^{2},\quad \forall i =1 , \cdots, m\\
\tilde{\mathbf{c}}_{ij} & = & \min_{\mathbf{c}_{ij} \in\mathbb{R}^{K}}\|a^{\xi}_{ij}-\sum_{k=1}^{K}c_{ij, k}f_{k}\|_{L_{\nu}^{2}}^{2},\quad \forall i,j =1 , \cdots, m
\end{eqnarray*}
\textit{ in the space $L_{\nu}^{2}$ of square-integrable functions
w.r.t. the measure $\nu$.}

\subsection{Learning of Effective Potentials}

In most physical applications, the dynamics of a system is determined by its
potential energy, which is a physically intuitive quantity.
For example, the overdamped Langevin dynamics in Eq.~(\ref{eq:overdamped_langevin})
is defined by the potential energy $U$, which generates the drift via its
gradient field. Learning the individual components of the drift separately as
in Eq.~(\ref{eq:regression_drift}) can pose a challenge in high dimensional
systems, since there is no guarantee the learned components are generated by a scalar potential.
To circumvent the problem, it is desirable to estimate the potential energy
directly instead of its gradient.

Let us consider the overdamped $d$ dimensional Langevin
dynamics Eq.~(\ref{eq:overdamped_langevin}).
We can introduce a dictionary of differentiable multivariate
basis functions $\Theta = \left( f_{1},\ldots,f_{K} \right)$, and make the
\emph{ansatz}:

\begin{eqnarray*}
U(x) & = & \sum_{k=1}^{K}c_{k}f_{k}(x) = \Theta(x) \cdot \mathbf{c},
\end{eqnarray*}

Define a tensor $\mathbb{D} \in\mathbb{R}^{d \times N \times K}$ and a matrix
$\mathbb{Y} \in\mathbb{R}^{d \times N}$ by
\begin{eqnarray}
D_{ilk} & = & \frac{\partial f_{k}}{\partial x_{i}}(X(t_{l})),\label{eq:definition_D_matrix}\\
Y_{il} & = & \frac{1}{s}(X_{i}(t_{l+1})-X_{i}(t_{l})).\label{eq:definition_Y}
\end{eqnarray}
and consider the following regression problem:
\begin{eqnarray*}
\tilde{\mathbf{c}} & = & \min_{\mathbf{c} \in\mathbb{R}^{K}}\|\mathbb{Y}-\mathbb{D} \cdot \mathbf{c}\|_{2}^{2}.
\end{eqnarray*}
where multiplication above represents summation over the last dimension of $\mathbb{D}$.
We show in appendix~\ref{sec:energy_relation} that the solution of such a
regression problem converges to the best approximation of the gradient field by
the linear combination $\nabla U(x)=\sum_{k=1}^K c_k \nabla f_k(x) = \nabla
\Theta(x) \cdot \mathbf{c}$, by construction. 

For a general diffusion process Eq.~(\ref{eq:ito_sde}), defining a generalized
potential that links drift and diffusion is still possible, if the dynamics are
reversible. In this case, there exists a scalar function, which we call
\emph{free energy} $\mathcal{F}:\mathbb{R}^{d} \mapsto \mathbb{R}$ such that~\cite{Pavliotis:2014aa}:
\begin{eqnarray}
\frac{\partial}{\partial x_{i}}\mathcal{F}(x) & = & \left[a^{-1}(x)(\frac{1}{\beta}\nabla\cdot a_{i}-b)\right]_{i}.
\label{eq:drift_diffusion_energy}
\end{eqnarray}
Here, we use the notation $\nabla\cdot a_{i}$ to denote the divergence
of the $i$-th row of the covariance matrix $a$. Eq.~(\ref{eq:drift_diffusion_energy})
also holds for the effective drift and diffusion $b^{\xi},\,a^{\xi}$
after applying a projection $\xi$, because the effective dynamics
discussed in the previous section inherits reversibility from the
original dynamics~\cite{Zhang_Faraday16}. Therefore, we discuss the
projected case in the following, as estimation of  the full dynamics
is a special case of this problem. 

Since the gradient of $\mathcal{F}$ in Eq.~(\ref{eq:drift_diffusion_energy})
now depends on two unknowns, we need to estimate one of them first before we
can solve for the free energy gradient. Suppose we have used the regression of
Eq.~(\ref{eq:regression_diffusion}) to obtain an expression for
each component of the diffusion matrix $a^{\xi}(x)$ as
\begin{eqnarray}
a_{ij}^{\xi}(x) & = & \sum_{k=1}^{K}c_{ij,k}f_{k}.
\label{eq:diffusion_basis}
\end{eqnarray}
This model allows to evaluate each component of the diffusion and
its derivatives at every simulation point. 

Next, we can use Eq.~(\ref{eq:drift_diffusion_energy}) and the convergence of
linear variations to the effective drift to set up a regression problem for the
free energy gradient as a linear combination of the vector fields $\nabla f_k$,
i.e.
\begin{eqnarray*}
\nabla \mathcal{F}=\sum_{k=1}^K v_k \nabla f_k.
\end{eqnarray*}
The regression problem becomes
\begin{eqnarray}
v & = & \min_{w\in\mathbb{R}^{K}}\|\mathbb{Y}-\mathbb{D} w\|_{2}^{2},
\label{eq:regression_energy}
\end{eqnarray}
where $\mathbb{Y}$ and $\mathbb{D}$ are now given by
\begin{eqnarray*}
Y_{il} & = & \left[\left(a^{\xi}\right)^{-1}(X(t_{l}))(\frac{1}{\beta}\nabla\cdot a_{i}^{\xi}(X(t_{l})) - e^{s}(X(t_{l+1}),X(t_{l})))\right]_{i},\\
D_{il,k} & = & \frac{\partial f_{k}}{\partial z_{i}}(X(t_{l})),
\end{eqnarray*}
and $e^{s}$ is a $m$-dimensional vector of finite differences,
\begin{eqnarray*}
e_{j}^{s}(y,x) & = & \frac{1}{s}(\xi_{j}(y)-\xi_{j}(x)).
\end{eqnarray*}
Please see Appendix~\ref{sec:energy_relation} for the detailed proof.


\section{Methods}

Our goal is to  find a sparse representation to the drift and diffusion term,
which requires computing a sparse solution to the regression problem
Eqs.~(\ref{eq:regression_drift}-\ref{eq:regression_diffusion}).
Standard regression can be biased towards sparse solutions by introducing a $L^{0}$ constraint (also known as \emph{subset selection})
into the standard optimization process
\begin{equation}
\tilde{\mathbf{c}}=\min_{\mathbf{c}} (\|\mathbb{Y}-\mathbb{X}\mathbf{c} \|^{2}_2+\rho \| \mathbf{c}\|_{0})
\label{eq:L0_norm}
\end{equation}
where the $0$-norm $\| \mathbf{c} \|_0$ denotes the number of non zero components in $\mathbf{c}$.
 
Unfortunately, the minimization problem Eq.~(\ref{eq:L0_norm}) is non-convex,
which makes finding a solution a NP-hard task. One popular way around this problem
is to relax the problem to a $L^1$-norm constraint, and a sparse solution can
then be computed by using one of the many algorithms available, such as
\emph{Lasso}~\cite{Tibshirani96Lasso, Tibshirani_perspective}, \emph{matching
pursuit}~\cite{matching_pursuit} and its orthogonal variant OMP~\cite{OMP}, or
\emph{elastic net}~\cite{ElasticNet}, just to name a few.
Independently of the specific protocol, the sparse solution
will have some coefficients equal to zero.

In principle, any sparsity value can be enforced in the solution, by
tuning the Lagrange multiplier $\rho$ in Eq.~(\ref{eq:L0_norm}). However, only a
subset of those values provide a representation of the data set that is both
accurate and compact. For instance, we expect an excessively sparse solution
$\mathbf{c}$ to severely under-fit and a barely sparse solution to over-fit the
data~\cite{hastie_book}. An under- (over-)fitted model contains less (more)
parameters than can be justified by that data, and both regimes should be avoided.
For this reason, any  algorithm enforcing sparsity needs to be complemented by a criterion
that allows to assess whether a solution is still statistically meaningful and
that signals whether the over-fitting or under-fitting regimes are entered, in
order to automatically select the sparsity level.  We propose to use the
statistical procedure of \emph{Cross Validation}~\cite{Kohavi95astudy} to
select solutions with optimal sparsity. 

We show in the following that when using Cross Validation sparsity can be
automatically enforced with iterative algorithmic formulations such as the
Stepwise Sparse Regressor. Despite its intrinsic simplicity and
intuitive interpretation, such an algorithm appears robust and effective, as it
is discussed below.

\subsection{Sparsity enforcement}

The approach we employ to solve the sparse regression Eq.~(\ref{eq:L0_norm}) for
stochastic systems is inspired by the iterative thresholding algorithm proposed
by Brunton \emph{et al.} in their deterministic SINDy
study~\cite{BruntonPNAS2016}, which works as follows. First, a standard
unconstrained linear regression is solved to compute a non-sparse solution
$\mathbf{c}$. Then, coefficients with a magnitude smaller than a pre-defined
threshold value $\lambda$ are set to zero and regression is performed on the
remaining coefficients, and the procedure is iterated till no coefficients are
found smaller than $\lambda$. The threshold parameter $\lambda$ is a
sparsification knob which needs to be tuned appropriately. 
While such an algorithm appears to produce good results in the identification
of deterministic differential equations from data, it is not robust for the
stochastic case. 

We modify the thresholding approach to enforce sparsity iteratively by
removing only one coefficient in every iteration, and use Cross Validation to
select the number of iterations, as it is discussed in the next section.
This modification removes the need of adjusting an external parameter like
$\lambda$. The pipeline works as follows:
\begin{itemize}
\item A standard least square regression $$\tilde{\mathbf{c}}=\min_{\mathbf{c}
\in \mathbb{R}^K} \|\mathbb{Y}-\mathbb{X}\mathbf{c} \|^{2}_2$$ is solved to
determine a preliminary (non-sparse) solution
$\tilde{\mathbf{c}}$.

\item One coefficient is set to zero. Different criteria can be used to select
the coefficient to remove, for instance the one with the smallest value can be
deterministically chosen, i.e.  
$$\tilde{c}_i = 0 : i = \min_k |\tilde{c}_k|$$ 
This way the level of sparsity is increased at every iteration.

\item Standard regression is performed again on the remaining degrees of
freedom $$\mathbb{Y} = \mathbb{X}[:, \hat{i}]
\tilde{\mathbf{c}}[\hat{i}]$$ where $\hat{i}$ indicates the set of all
dictionary indexes but $i$, which has been removed.

\item The procedure is iteratively repeated until Cross Validation indicates
that the optimal sparsity level (i.e. number of iterations) $\tilde{s}$ in
the solution $\tilde{\mathbf{c}}$ is reached.
\end{itemize}

We call this algorithm \emph{Stepwise Sparse Regressor} (SSR), and
introduce the shorthand notation
\begin{equation}
SSR(\mathbb{X}, \mathbb{Y})_{k}
\label{eq:SSR_shorthand}
\end{equation}
to indicate the solution $\mathbf{c}$ obtained upon running the algorithm on $k$
iterations. 
Such a solution is $k$-sparse, e.g. has $k$ zero coefficients and $n = K - k$
non-zero coefficients. In the following, we are going to refer to the parameter $n$ as the 
\emph{solution size}, while discussing the results.

Once Cross Validation is used to identify the number of iterations
corresponding to the optimally sparse solution, the algorithm is
parameter free and does not require any preliminary training phase before use.

\subsection{Cross Validation}

The specific number of iterations on which the Stepwise Sparse Regression needs
to be run to find the optimal solution is determined by a \emph{Cross
Validation} (CV) calculation~\cite{geisser1975predictive},  a statistical validation
technique that has risen to great popularity in the interdisciplinary fields of
model and hyperparameter selections (see ref.~\citenum{hastie_book} for an
introductory self contained discussion). The underlying idea is straightforward
and summarized below. 

Let us assume we have a family of parametric models $(\mathcal{M}(\lambda_1),
\cdots , \mathcal{M}(\lambda_r))$  depending on a hyperparameter $\lambda$ which
takes values $\lambda_1, \cdots, \lambda_r$ and we would like to select the one
model that fits best a given data set $\mathcal{D}$. In the original CV
formulation, the full data set is split into two disjoint subsets, and each
model in the family is alternatively trained on one of them first and then
tested on the other. The  \emph{cross validation score} is the average
deviation $\delta$ of the predictions of the trained model from the actual test
set, and it measures how accuracy and predictivity are balanced in that model. 
The set of parameters yielding low values of $\delta$ are selected and
identify ''optimal'' models in the family.

Here, we use CV to select the size $n$ of the optimal solution to
the linear regression problem Eq.~(\ref{eq:L0_norm}), which plays the role of the
hyperparameter $\lambda$ from the last paragraph. The family of models to
validate is now a set of SSRs with different solution size $n = K - q$ (or,
equivalently, different sparsity $q$), i.e.
$$\left\{  SSR_q\right\}_{q=1, \cdots, K},$$ 
where the notation introduced in the previous section is used. CV is run on
each model to generate a family of cross validation scores $\delta[SSR_q]$. 

We use a $k$-fold cross validation formulation, where the full
dataset is split into $k$ subsets, each of them playing alternatively the role of
test set in a $k$ step procedure. Let us start by partitioning the dataset
$\mathcal{D}$ containing $N$ data points $p$ into $k$
disjoint equivalent subsets $A_i$, which are selected randomly, i.e.
$\bigcup_i A_i = \mathcal{D}, \quad A_i \cap A_j = \emptyset$. Moreover, let us
introduce the shorthand: 
$$\mathbb{X}_{A_i} = \mathbb{X}[p_{A_i},:], \quad p_{A_i}  = \bigcup_{p \in A_i} p$$ 
Then, the cross validation score for each model $SSR_q$ is defined as an
average
\begin{align}
\delta^2[SSR_q] = \frac{1}{k}\sum_{i=1}^{k}  \|    \mathbb{Y}_{A_i} -  & 
\mathbb{X}_{A_i} \cdot SSR(\mathbb{X}_{B_i}, \mathbb{Y}_{B_i})_q      \| ^2_2 \\
& B_p = \bigcup_{i \neq p} A_i
\label{eq:CV_score}
\end{align}
where $SSR(\mathbb{X}_{B_i}, \mathbb{Y}_{B_i})_{q}$ indicates the $q$-sparse
linear combination coefficients generated by running \emph{SSR} on the training
set $B_i$ (see notation Eq.~(\ref{eq:SSR_shorthand})), which are then used to make a prediction $\mathbb{X}_{A_i} \cdot
SSR(\mathbb{X}_{B_i}, \mathbb{Y}_{B_i})_q$.
\begin{figure}[h]
\centering{}
\includegraphics[scale=0.3]{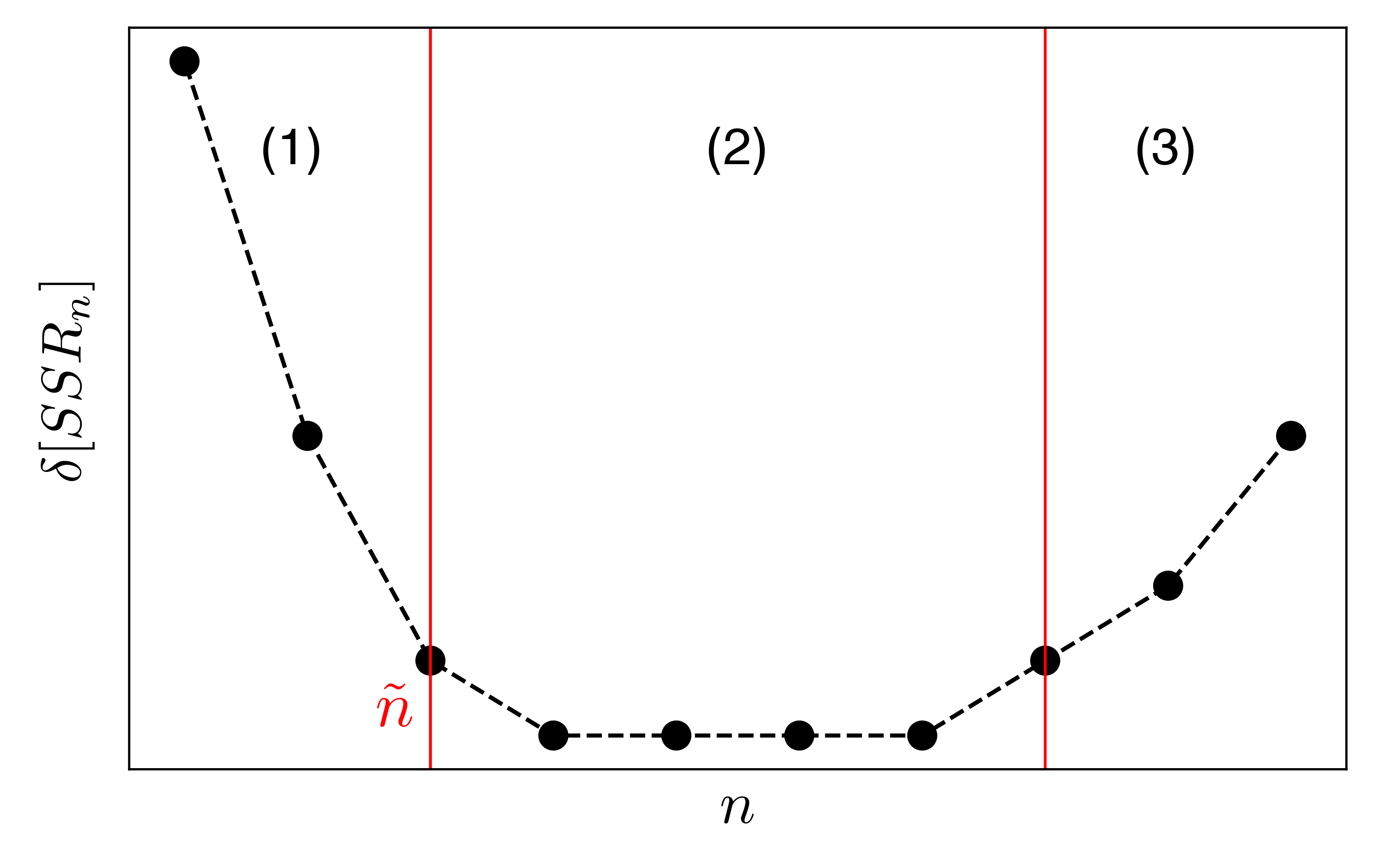}
\caption{Cartoon representation of the expected behavior of the $k$ fold cross
validation score $\delta[SSR_n]$ as a function of the solution size
$n$  in the linear combination solution. \label{fig:CV_cartoon}}
\end{figure}
The set of $\delta[SSR_q]$ is then monitored as a function of the solution
size $n = K - q$, which usually results in a a behavior close to that represented in
Fig.~\ref{fig:CV_cartoon}. 
We expect an intermediate regime of low cross validation score (accurate)
solutions (region (2)) with variable sparsity: all such solutions are equally
good at balancing sparsity and accuracy.  In addition, this regime is bounded
from the right and left by and under- and over-fitting regime (regions (1) and
(3)) respectively . Solutions belonging to  both regimes are characterized by
larger values of the cross validation scores, which indicate that accuracy is decreasing.
Intuitively, the one solution separating regime (2) from regime (1) is what we
call optimally sparse $\tilde{n}$, since:
$$\frac{\delta[\tilde{n} - 1]}{\delta[\tilde{n}]} \gg 1, \quad
\frac{\delta[\tilde{n}]}{\delta[\tilde{n}+1]} \approx 1 $$ 
The $\tilde{n} \rightarrow \tilde{n}+1$
gap in the $\delta$ values is a clear signal that increasing sparsity by one
additional unit compromises the model predictive power.

In the following, the optimally sparse solution $\tilde{n}$ is chosen by identifying such a
transition point in the cross validation score curves.
All cross validation calculations reported below were performed using Python
routines available in \emph{sklearn}~\cite{scikit-learn}.

\section{Homogeneous diffusion in a double well potential}

\begin{figure}[h]
\centering{}
\includegraphics[scale=0.6]{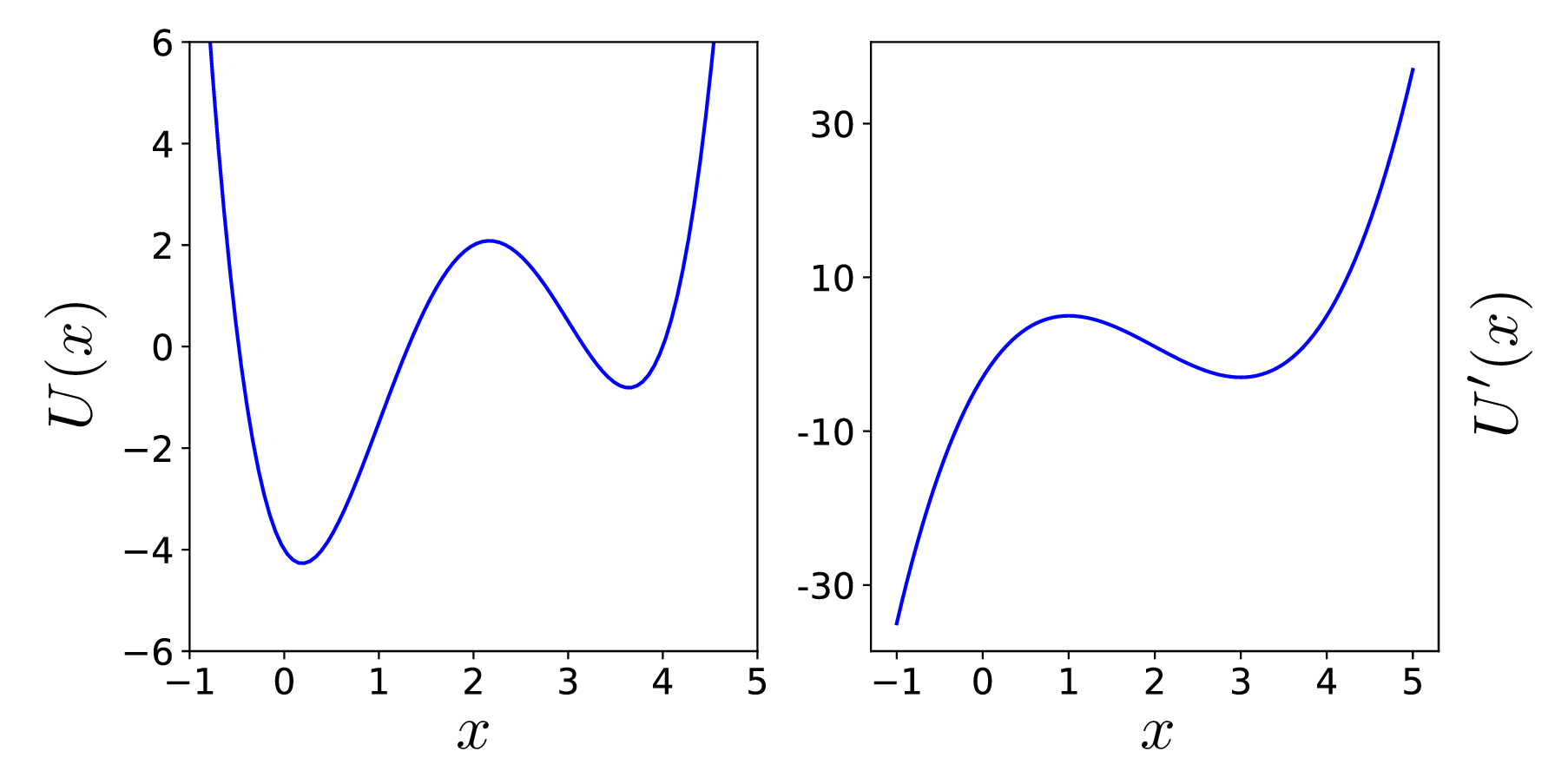}
\caption{Potential energy profile $U(x)$ (left panel) and its gradient
$U^{\prime}(x)$, which equals the opposite of the force  (right panel).
\label{fig:1d_profile}}
\end{figure}
We illustrate our sparse regression protocol by applying it to overdamped
homogeneous diffusion data in a one dimensional double-well polynomial
potential Fig.~\ref{fig:1d_profile}:
\begin{equation}
U(x)=\frac{1}{2}x^{4}-4x^{3}+9x^{2}-3x,\qquad\frac{dU}{dx}=2x^{3}-12x^{2}+18x-3
\label{eq:pot1}
\end{equation}
to recover the drift of the process from the data.
Five independent trajectories were generated by integrating
Eq.~(\ref{eq:overdamped_langevin}) using the gradient in Eq.~(\ref{eq:pot1}).
We computed the time increments Eq.~(\ref{eq:simple_increm}) and averaged them
over discrete bins as explained in Appendix~\ref{sec:binning}. We illustrate
the performance of the sparse
regression on two different dictionaries $\Theta$ and $\Theta^\prime$. Each
dictionary consists of $K = 20$ basis functions of the form $[1, x, x^2,
x^3, \cdots]$. Details about the dataset and the specific composition of the two
dictionaries are provided in the Appendix, Eqs.~(\ref{eq:theta}) and (\ref{eq:thetaprime}).
The first four entries in both databases correspond to
the functions composing the drift term used to generate the data, $[1, x, x^2,
x^3]$. We refer to the these four functions as \emph{analytic basis
functions} in the following.

\begin{figure*}
\begin{centering}
  \includegraphics[width=0.9\textwidth]{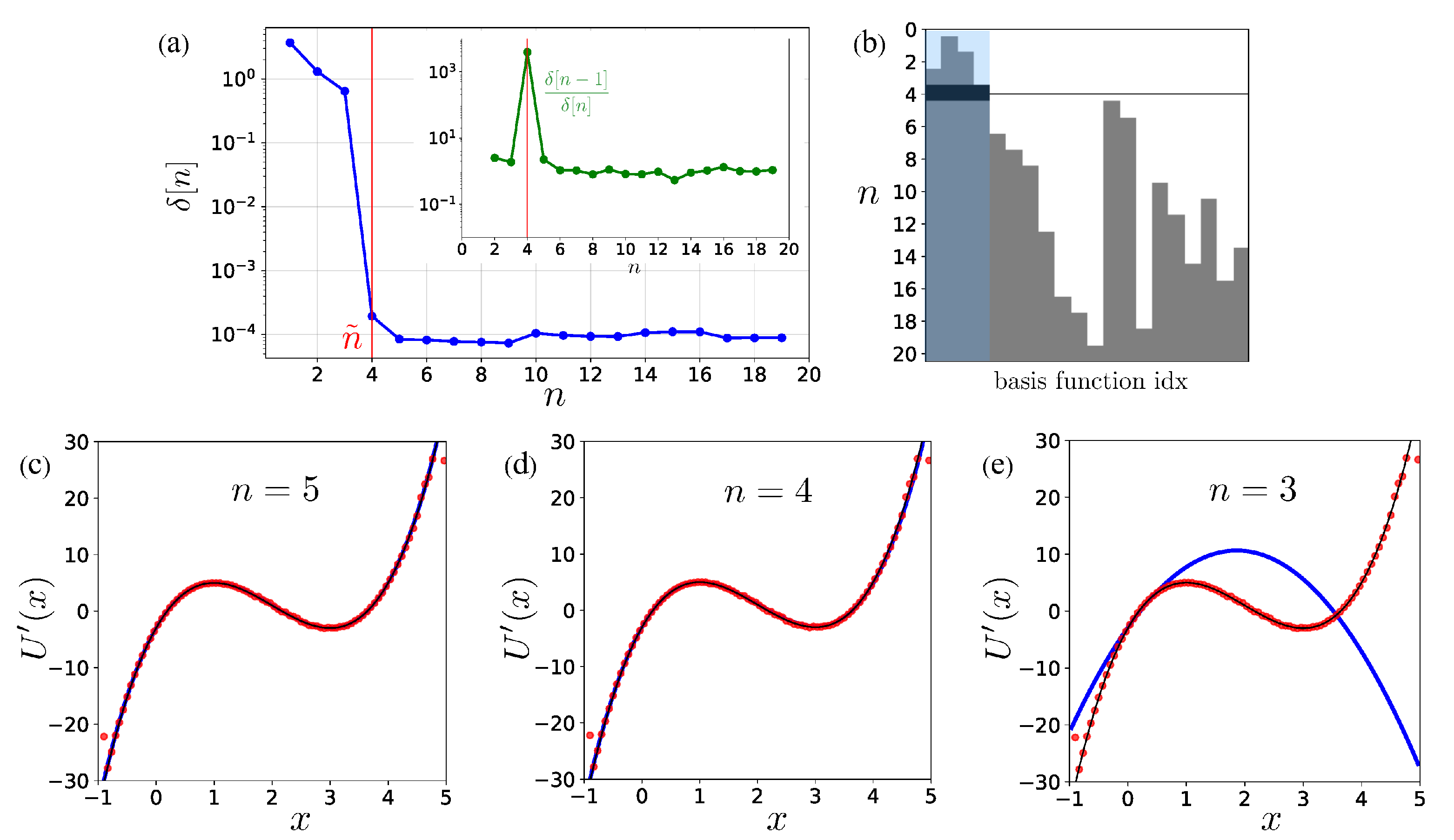}
  \caption{
  Results from applying the  \emph{SSR} algorithm to a trajectory
  generated by diffusion in the potential of Eq.~(\ref{eq:pot1}), using a function dictionary
  $\Theta$. (Panel (a)) Cross validation score is plotted as a
  function of solution size $n$.  
  The inset shows the ratio $\delta[n-1]/\delta[n]$ as a function of $n$. Vertical red lines indicate the 
  number of non zero coefficients $\tilde{n} = 4$ in the optimal solution.
  (Panel (b)) Sparsity progress matrix: any $(i,j)$  (grey) white
  entry indicates that coefficient $c_i$ is (non) zero after $K-j$ ($K=20$) iterations.
  A horizontal black line indicates $\tilde{n}$, and the corresponding coefficients are
  colored in black. The four dictionary basis functions $[1, x, x^2, x^3]$ have
  survived, and are highlighted in a blue color. 
  (Panels (c)-(e)) Comparison
  between the exact gradient $U^\prime (x)$ (black solid line) and
  solutions $\tilde{U}^\prime(x)  = \sum_k \tilde{c}_k f_k(x)$  (blue solid
  lines) with decreasing solution size $n$ values  (or increasing sparsity, from left to right). Red
  markers represent the binned time increments as from Eq.~(\ref{eq:Kramers_Moyal_1}).
  \label{fig:comparison_1d}}
  \end{centering}
\end{figure*}

Fig.~\ref{fig:comparison_1d}a shows the cross validation
score $\delta_{\Theta}$ from Eq.~(\ref{eq:CV_score}) for the first dictionary,
$\Theta$,   as a function of the solution size $n$. As the number of surviving
coefficients $n$ decreases (from right to
left), the cross validation score stays constant; however, going from a four
term $n = 4$ to a three term $n = 3$ solution causes the cross validation
score to increase by several orders of magnitude, which suggests that sparsity
is now too extreme and compromises the model predictivity. The signal is clear, as indicated by the 
plot of ratios $\delta[n-1] / \delta[n]$  as a function of $n$ (inset in Fig.~\ref{fig:comparison_1d}a).
The position of the gap in the CV score curve suggests that $n = 4$ is the
optimal solution sparsity, which is associated with a $\tilde{\delta}_{\Theta} =
1.49 \cdot 10^{-4}$ CV score.

Fig.~\ref{fig:comparison_1d}b shows the sparsity progress matrix, 
which monitors the linear combination status as a function of sparsity.
The $(i,j)$ entry in the matrix refers to the function $f_i$ in the dictionary,
when the solution only contains $n = j$ terms (i.e. after $20-n$ iterations).
The color code is as follows: grey pixels
indicate that the coefficient $c_i$ is still alive, whereas white pixels are
used for coefficients that have been removed. A horizontal black line indicates
the optimal solution size $\tilde{n}$, as from the cross validation score plot,
and the corresponding pixels are also colored in black. A light blue color is
used to mark the analytic basis functions. Please note that no coefficient is
resuscitated during the SSR iterations after it is removed from the
dictionary in a previous iteration.

The optimal solution only contains the analytic basis functions and reads:
\begin{equation}
\tilde{U}^\prime_{\Theta} (x)  = \tilde{\mathbf{c}} \cdot \Theta(x) = -2.98 + 17.84 x - 11.82  x^2 + 1.96 x^3
\label{eq:1d_solution_ok}
\end{equation}
This expansion is an accurate approximation to Eq.~(\ref{eq:pot1}), as it can be
seen by comparing coefficients and from panel d in Fig.~\ref{fig:comparison_1d}. 

It is instructive to compare the optimal solution
Eq.~(\ref{eq:1d_solution_ok}) to a slightly less sparse $n = 5$ and to a sparser
$n = 3$ solution. The explicit expansions for these cases are: 
\begin{align}
U^\prime_{n = 5}(x)  & = -2.98 + 17.43 x - 11.75  x^2 + 1.98 x^3 + 0.4 \sin x \\
U^\prime_{n = 3}(x)  & = -2.88 + 14.5 x - 3.89  x^2
\end{align}
and are plotted together with the gradient Eq.~(\ref{eq:pot1})
in Fig.~\ref{fig:comparison_1d}c and Fig.~\ref{fig:comparison_1d}e. 
The $n = 5$ solution contains the analytic functions and an extra small oscillatory
term, and accurately approximates the gradient, $\delta = 7.8 \cdot 10^{-5}$.
In contrast, the sparser solution $n=3$ is deprived of one key dictionary
ingredient and does not perform well, as shown by panel e of
Fig.~\ref{fig:comparison_1d} and the much larger cross validation score $\delta=9
\cdot 10^{-1}$. 

\begin{figure*}
\begin{centering}
  \includegraphics[width=0.9\textwidth]{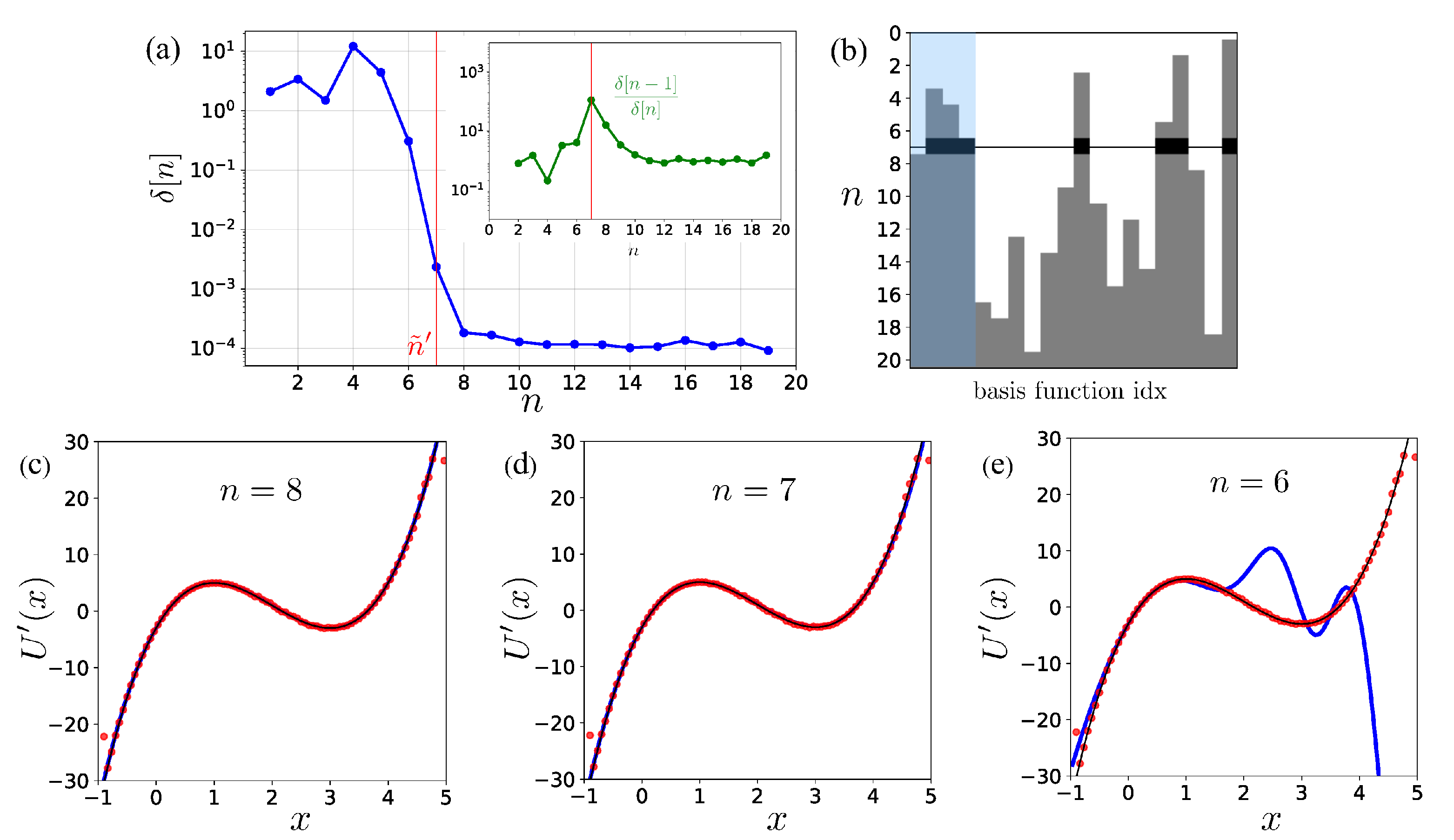}
  \caption{
  Sparse regression results for a dictionary $\Theta^\prime \neq \Theta$,
  using the same notation and color code as in Fig.~\ref{fig:comparison_1d}.
  (Panel (a)) Cross validation score as a function of solution size $n$
  The inset shows the ratio $\delta[n-1]/\delta[n]$ as a function of $n$. Vertical red lines indicate the 
  number of non zero coefficients $\tilde{n}^\prime = 7$ in the optimal
  solution. (Panel (b)) Sparsity progress matrix. (Panels (c)-(e)) Comparison
  between the exact gradient $U^\prime (x)$ (black solid line) and
  solutions $\tilde{U}^\prime(x)  = \sum_k \tilde{c}^\prime_k \Theta^\prime(x)$ (blue solid
  lines) with decreasing solution size $n$ (or increasing sparsity, from left to right). 
  Red markers represent the binned time increments as from Eq.~(\ref{eq:Kramers_Moyal_1}) 
  Even though $\tilde{c} \neq \tilde{c}^\prime$, the gradient is still accurately approximated.
  \label{fig:degeneracy_1d}}
  \end{centering}
\end{figure*}

Let us now discuss the solution to the problem Eq.~(\ref{eq:Kramers_Moyal_1}) when a
different dictionary $\Theta^\prime$ is used. Results are summarized in
Fig.~\ref{fig:degeneracy_1d}, using the same format as in
Fig.~\ref{fig:comparison_1d}. 

The cross validation score plot in Fig.~\ref{fig:degeneracy_1d}a shows a similar trend as in
Fig.~\ref{fig:comparison_1d}a, but a clear gap in the $\delta$ values is now
missing, and suggests that an optimal solution is somewhere in the range
between $n = 5$ and $n = 10$. The transition point 
$\tilde{n}^\prime_k = 7$ is selected as the value of $n$ maximizing
the CV score ratio $\delta[n-1]/ \delta[n]$, as shown
in the inset. It is worth noting that the maximum amplitude of the ratio is
here two orders of magnitude smaller than in Fig.~\ref{fig:comparison_1d}a.
The progress matrix  in Fig.~\ref{fig:comparison_1d}b shows that
the three analytic terms $[x, x^2, x^3]$ are present in the optimal solution, but
$f_0 = 1$ is not (first column), and there are additional contributions. 
The actual expansion $\tilde{U}^\prime (x) = \tilde{\mathbf{c}}^\prime \Theta^\prime (x)$ reads:
\begin{align}
\begin{split}
\tilde{U}^\prime (x)   = & 17.85 x -12.68 x^2 + 2.11 x^3 + 9.61\exp \left(-50(x-3)^2 \right) + \\ 
&  -2.97\exp \left(-50 (x-4)^2\right) -18.77\exp \left(-0.6 (x-4)^2\right) + \\ 
&   + 12.97 \left[ \tanh^2(x-4) +1 \right]
\label{eq:suboptimal}
\end{split}
\end{align}

Performing a CV score based SSR on the two different dictionaries produces two
solutions with different levels of sparsity and cross validation scores. Both
of them succeed at capturing the double well feature
Fig.~\ref{fig:comparison_1d}c and Fig.~\ref{fig:degeneracy_1d}c of the potential
Eq.~(\ref{eq:pot1}). As a matter of fact, a Taylor expansion of the $\Theta^\prime$-solution 
Eq.~(\ref{eq:suboptimal}) returns a polynomial series which is consistent with 
Eq.~(\ref{eq:1d_solution_ok}).

However, the solution associated with dictionary $\Theta^\prime$ is
less parsimonious than what was found for the dictionary $\Theta$ considered above,
and it is associated with a larger cross validation score
$$\tilde{\delta}_{\Theta} = 1.49 \cdot 10^{-4} <
\tilde{\delta}_{\Theta^\prime}= 5.6 \cdot 10^{-3}.$$
For this reason, the solution for dictionary $\Theta$ should be favored over
$\Theta^\prime$.
We show in the next section that even if different dictionaries return different
optimal solutions, the comparison of results for different
dictionaries leads to the identification of the maximally sparse solution.

\subsection{Greedy search}
The results discussed above indicate that the performance of the SSR algorithm
is affected by the composition of the dictionary used. The resulting 
optimal solutions are quantitatively different but qualitatively very
similar to one another.
This section is devoted to investigating this issue systematically.

The comparison of Fig.~\ref{fig:comparison_1d} and Fig.~\ref{fig:degeneracy_1d}
shows that the cross validation signature when the SSR identifies the
correct (maximally sparse) solution is much stronger than for the case of a
less sparse solution.

This consideration suggests that, instead of proceeding by iteratively removing
functions from the dictionary, cross validation could be used to extensively
test all possible combinations of basis functions and compare the results to
determine the maximally sparse solution.
That is, given a large reference dictionary, all possible combinations of
functions (with a given sparsity) could be considered, and the corresponding
CV-score estimated and compared.

We illustrate this idea by defining a large reference dictionary $\Omega$ of
$M=100$ basis functions. The two previously used dictionaries are
included in this large one $\Theta, \Theta^\prime \subset \Omega$.
The standard (non-sparse) linear regression problem
$$\mathbf{c}_\theta = \min_{\mathbf{c}}\| \mathbb{Y} - \mathbb{X}_{\theta} \mathbf{c}\|^2_2$$ 
can be solved for each sub-dictionary $\theta \subset \Omega$ with $n < M$
functions, and the associated cross validation score $\delta_{\theta}$ can be
computed. If the $\delta_{\theta}$ values for all possible sub-dictionaries
$\theta \subset \Omega$ and different $n$ values are tabulated, the optimal
solution can be identified by comparing the change in cross validation as a
function of the dictionary size.
Such a procedure evaluates and compares different levels of sparsity in a greedy
fashion and can in principle replace the need of an iterative sparse regression
algorithm.

For a very large reference database of $M$ functions, such a brute force
approach becomes computationally very demanding as the number of different
sub-dictionaries with $n$ functions is given by the binomial coefficient
$C_{M,n} = \binom{M}{n}$, which becomes untreatable if $n \gg 1$.
In order to demonstrate the approach, we randomly sample $m_n$ different
$n$-function dictionaries $\theta$ from $\Omega$, for
increasing values of $n$, and run CV validation on each of them (details on
specific values of $m_n$ are given in the Appendix). 

The main results are shown in Fig.~\ref{fig:CV_greedy}, where the
cross validation scores $\delta_{\theta}$ (averaged over several realizations,
as detailed in the Appendix) are plotted as a function of the dictionary size $n$.
\begin{figure}[h]
\centering{}
\includegraphics[scale=0.4]{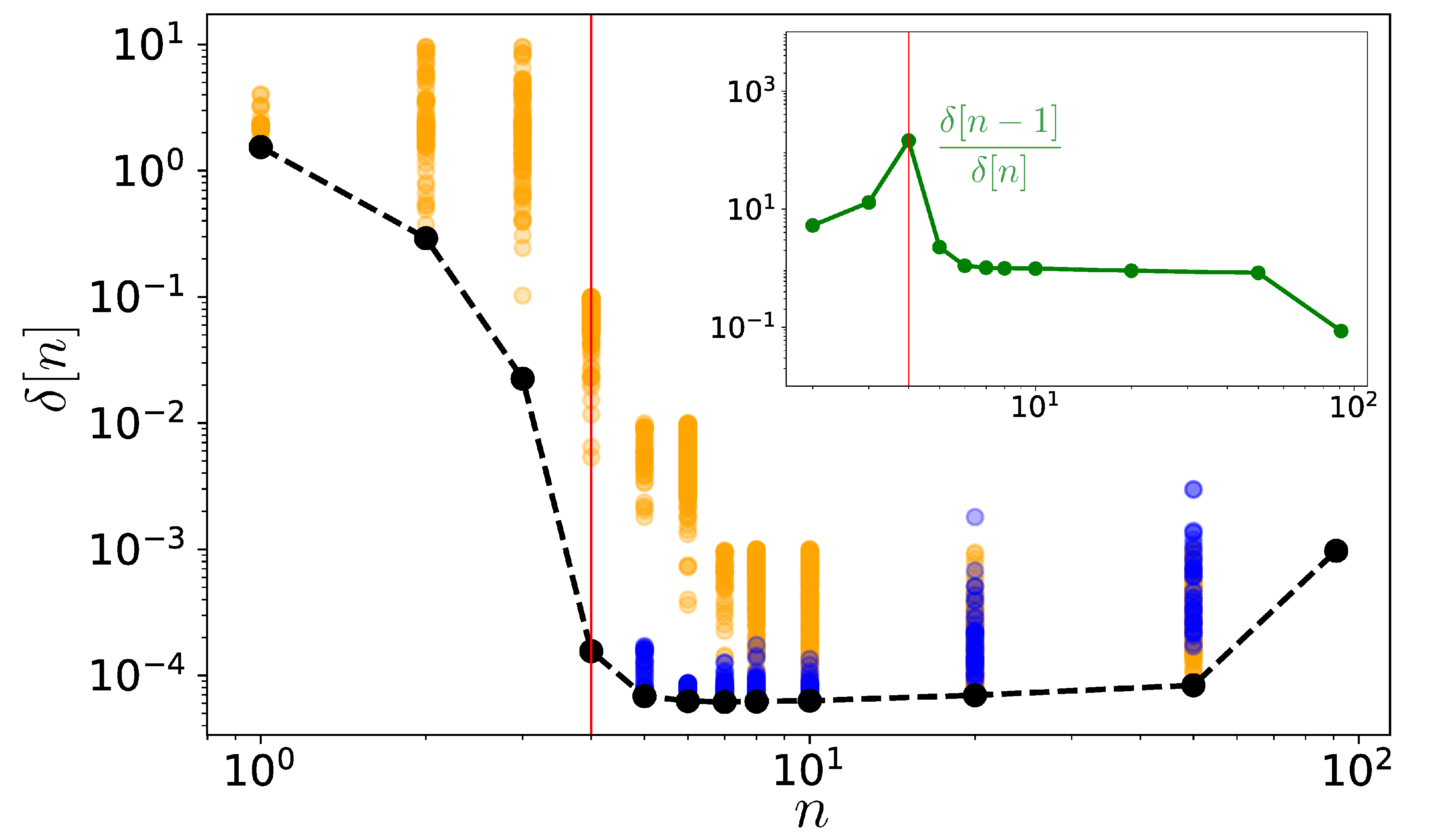}
\caption{Greedy search in the solution space to the regression problem,
Eq.~(\ref{eq:L0_norm}), applied to the two-well potential, Eq.~(\ref{eq:pot1}).
For a given size $n$, the cross validation score for a set of $100$
different dictionaries $\delta_{\theta, n}$ (subsets of the large reference
dictionary $\Omega$) are plotted. Each marker represents a dictionary instance, and
is colored in blue if the dictionary contains the analytic basis functions  $[1,
x, x^2, x^3] \subseteq \theta$, or in orange otherwise.
The minimal (optimal) cross validation score $\min(\delta_k)$ over all
realizations with a given size $n$ is indicated with a black dashed line.
A vertical red line indicates the solution with optimal sparsity, as
determined by the peak in the function $\delta[n-1]/\delta[n]$, as shown in the
inset and represents the optimally sparse solution $\tilde{\theta} = [1, x,
x^2, x^3]$. \label{fig:CV_greedy}}
\end{figure}

Each dictionary realization is described by an orange point $(n, \delta_\theta)$,
unless the analytic basis functions $[1, x, x^2, x^3] \subset \theta_n$, in
which case the point is blue. The minimum cross validation score
was selected (marked in black in Fig.~\ref{fig:CV_greedy}) across all points
for each dictionary size $n$, and the resulting curve is
plotted using a dashed black line.

Fig.~\ref{fig:CV_greedy} shows that the optimal sparse solution
$\tilde{\theta}=[1, x, x^2, x^3]$, which is associated with $\delta
= 1.49 \cdot 10^{-4}$ (as also obtained for the $\Theta$-solution in the last
section), clearly represents the transition point in the cross validation
curve. Any other $n=4$ dictionary has a larger cross validation score,
as signaled by the gap between the blue point and all other orange
realizations at $n=4$. Moreover, any sparser ($n=3$) dictionary has an associated
cross validation score larger by several orders of magnitude. The increase in
CV score is a footprint that a dictionary is missing (at least) a key component
(as in the cross validation score plots in Figs.\ref{fig:comparison_1d} and 
Fig.~\ref{fig:degeneracy_1d}). As already observed before,
less sparse solutions such as $5 < n <20$ may
reproduce the gradient of the potential equally good or slightly better and
they all have comparable cross validation scores. When $n \gg 1$ (as the
size of the dictionary increases), the cross validation scores start
increasing, indicating the over-fitting regime.

The $\Theta^\prime$ solution that was found in Fig.~\ref{fig:degeneracy_1d} is
represented by a $n=7$ orange point in Fig.~\ref{fig:CV_greedy}, together with
many others that share the same sparsity.

This greedy analysis shows that cross validation identifies the analytic
basis functions $\tilde{\theta} = [1, x, x^2, x^3]$ as the optimal sparse
solution to the problem Eq.~(\ref{eq:regression_drift}); sparser solution are
less accurate (larger $\delta$), and comparably accurate solutions are less
parsimonious (reduced sparsity), as indicated also by the plot of  $\delta[n-1]/\delta[n]$
as a function of solution size $n$ (inset in Fig.~\ref{fig:CV_greedy}). 

This analysis calls attention to the shortcoming of the deterministic SSR 
algorithm to search for the optimal solution. As seen in the previous section the 
performance of SSR depends on the choice of the database. The deterministic
nature of SSR is not always efficient in searching the solution space of the
non-convex problem and SSR can be trapped in local minima that provide a
sub-optimal solution.

The performance of the SSR algorithm can not be \emph{a priori} estimated by
considering indicators of the ill-conditioning of the dictionary. For the example
of the two dictionaries $\Theta$ and $\Theta^\prime$ used in the previous section
the condition number $\kappa$ of the database does not reflect their
performance, as $$\kappa(\mathbb{X}_{\Theta}) \approx 10^7, \qquad
\kappa(\mathbb{X}_{\Theta^{\prime}}) \approx 10^3.$$

At the level presented here, the proposed CV-based SSR is effective at
relaxing to a sparse (even if not always the sparsest) solution which is
dictionary dependent but which still efficiently captures the main features of
the gradient.

\subsection{Effects of sampling noise on algorithmic performance}

In this section we investigate to what extent the convergence of the sparse
regression algorithm is influenced by the presence of noise in the stochastic
system, by using the double well potential Eq.~(\ref{eq:pot1}) as a reference
system. One of the main assumptions underlying the proof of Theorem 1 is that
the data points are Boltzmann distributed. However, this condition is met only
approximately on a finite size trajectory. 

In order to investigate how deviations from the Boltzmann distribution affect
the performance of the algorithm, we bin the trajectory along the x-axis in $Q$
bins and introduce the bin-dependent relative error: 
\begin{equation}
\epsilon_{i}=\left|\frac{Y_{i}-U^\prime(\bar{x}_i)}{U^\prime(\bar{x}_i)}\right|
\label{eq:err}
\end{equation}
where $Y_i$ is the average of the time increments Eq.~(\ref{eq:simple_increm})
associated with bin $i$, and  $\bar{x}_i$ is the bin center, see also Eq.~(\ref{eq:binning}). The median of the error
distribution $\left\{ \epsilon_i \right\}_{i=1, \cdots, Q}$ over the trajectory
sample used in the previous sections is approximately $10^{-2}$ and the largest
deviation is found for the bin located on the top of the energy barrier (see
Fig.~\ref{fig:1d_profile}). 
The error could be decreased by running longer trajectories or lowering the
temperature of the system. Here we mimic these effects by generating new
samples $\hat{Y}_i$ in every bin $i$ with increasingly smaller deviation from
the ideal sample, using a scaling factor $f<1$:
\begin{equation}
\hat{Y}^f_{i}=U^\prime(\bar{x}_i) \mathcal{N}(0, \zeta), \qquad \zeta = f \cdot \mathrm{median}(\epsilon).
\label{eq:noise}
\end{equation}
The noise distribution $\mathcal{N}$ was chosen to be Gaussian to approximate
the effect of thermal noise. 
In practice the same effect could be obtained by running umbrella sampling
simulations in every bin. The sparse regression was then run on
this new, less noisy data set, for $100$ different random $50$ basis function
dictionaries $\Theta$ (all including $[1, x, x^2, x^3]$ entries) and different values
of the scaling parameter $f$. The performance of the SSR algorithm as a
function of the sampling error is reported in Tab.~\ref{tab:noise}.
\begin{center}
\begin{table}
\begin{tabular}{c | c c c c c}		
  $\log_{10}f$ & $0$ & $-3$ & $-6$ & $-9$ & $-\infty$ \\
  \hline
  $\%$ & 0 & 0 & 46 & 81 & 94
  \end{tabular}
  \caption{Percentage of dictionaries for which SSR converges to the optimal
  solution $[1, x, x^2, x^3]$, as a function of the noise scaling factor $f$,
  Eq.~(\ref{eq:err}). $f = -\infty$ is a shorthand indicating no
  sampling noise.}
 \label{tab:noise}
 \end{table}
\end{center}
While for large sampling errors the SSR algorithm converges to
the optimal solution $[1, x, x^2, x^3]$ for none of the $100$ randomly selected
dictionaries, the percentage of dictionaries where the optimal solution is found by the
SSR increases as the sampling noise is
reduced. Surprisingly, a handful of dictionaries still
relax to a sub-optimal minimum even in the absence of noise. 

This result indicates that for stochastic systems the sampling quality plays an
important role in determining whether the SSR algorithm gets trapped in a local
minimum in the solution space.  Additionally, the effect of the noise reduction
is dictionary-specific, as some dictionaries do not reach the global minimum
till extremely low level of noise.

In practice, relevant (usually high-dimensional) systems will present large
sampling errors, and a more robust algorithm for the search in the solution
space is needed.

\section{``Learning'' a projected dynamics}
\begin{figure}[!ht]
\begin{centering}
  \includegraphics[scale=0.6]{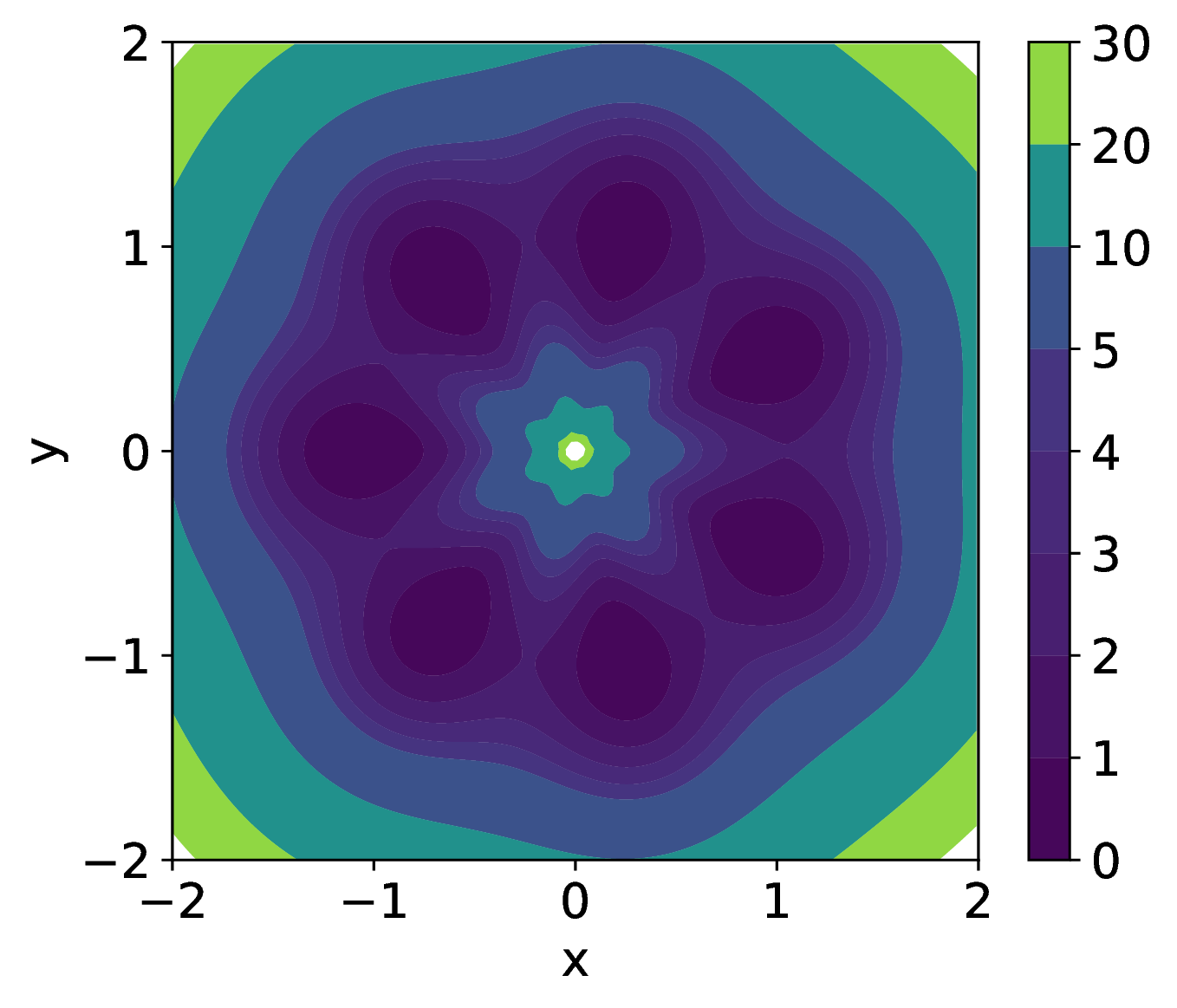}
  \caption{A contour plot $U = U(x,y)$ of the lemon slice potential
  Eq.~(\ref{eq:lemon_slice}) is shown. Seven angular mimima $\kappa=7$ can be
  clearly distinguished. Details are given in the text.
  \label{fig:contour}}
  \end{centering}
\end{figure}

\begin{figure*}[!ht]
\begin{centering}
  \includegraphics[width=\textwidth]{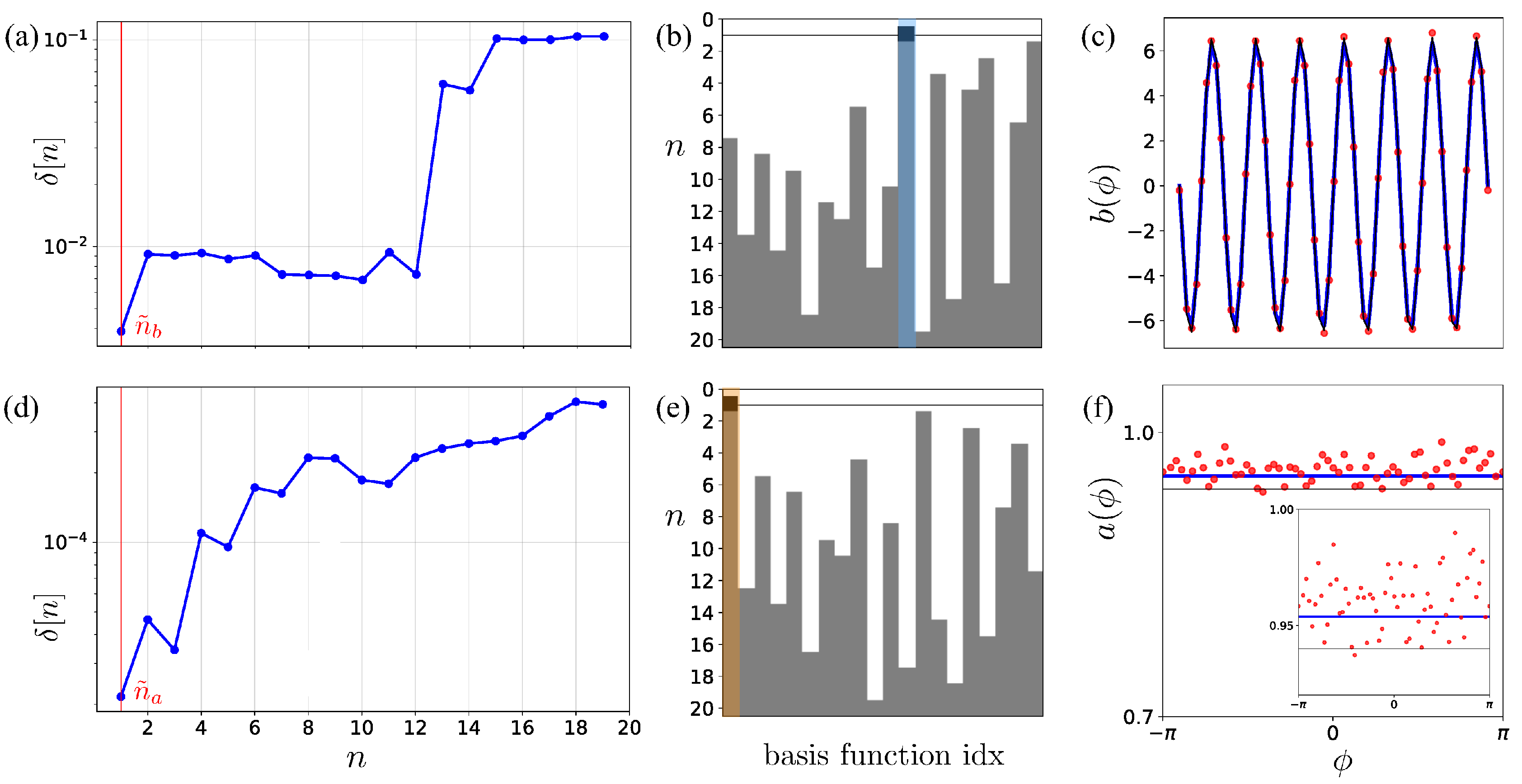}
  \caption{
  Sparse regression results for both drift $b$ (upper panels),
  and diffusion $a$ (lower panels) are summarized, using the same
  notation and color code as in Fig.~\ref{fig:comparison_1d}. (Panels (a) and (d))
  Cross validation score $\delta$ as a function of the solution size $n$,
  with vertical red lines indicating the optimal solutions $\tilde{n}_b$ and $\tilde{n}_a$.
  (Panels (b) and (e)) Sparsity progress matrices, where entries associated to the
  analytic functions are highlighted with color. (Panels (c) and (f)) Comparison between actual drift (diffusion) as from
  Eq.~(\ref{eq:eff_dyn}) (black line) and the optimal sparse representations $\tilde{b}(\phi)$
  ($\tilde{a}(\phi)$) obtained with the SSR algorithm (blue). Red markers are used to indicate estimation from sampling according to
  Eqs.~(\ref{eq:Kramers_Moyal_eff_1}-\ref{eq:Kramers_Moyal_eff_2}).
  \label{fig:comparison_lemon}}
  \end{centering}
\end{figure*}
We apply the analysis protocol discussed in the previous sections to ''learn''
the sparse stochastic dynamics along an effective
coordinate~\cite{Legoll_Nonlinearity2010}. As a benchmark, we use a system where the solution can
be computed analytically, that is, the two dimensional \emph{lemon slice}
potential introduced by Bittracher \emph{et al.}~\cite{Bittracher2017}, which
is specified by the polar representation $\left( \theta, r \right)$:
\begin{equation}
U(r,\phi) = \cos \left( \kappa \phi\right) + 10(r-1)^2 + \frac{1}{r}
\label{eq:lemon_slice}
\end{equation}
where $\kappa=7$ indicates the number of minima in the energy landscape, as shown in
Fig.~\ref{fig:contour}.

It was previously shown~\cite{Bittracher2017} that the polar angle $\phi \in [-\pi, \pi]$
correlates with the first seven eigenvectors of the backward Fokker Planck
operator associated with this potential, which describe the
basin hopping motions. The polar angle $\phi$ is then a good candidate for an
effective coordinate. Projecting the overdamped diffusion into this coordinate
results in the projected dynamics:
\begin{equation}
\begin{gathered}
d\phi_t = b(\phi)dt + \sqrt{2\beta^{-1} a(\phi)} d\eta(t) \\
b(\phi) = \frac{C_r}{Z_r}\kappa \sin \kappa \phi, \qquad a(\phi) = \frac{C_r}{Z_r}
\label{eq:eff_dyn}
\end{gathered}
\end{equation}
with the constants:
\begin{equation}
\begin{gathered}
C_r = \int_0^\infty r^{-1}\exp \left[ -10\beta (r-1)^2 - \beta/r\right]dr \\
Z_z = \int_0^\infty r\exp \left[ -10\beta (r-1)^2 - \beta/r\right]dr \\
\frac{C_r}{Z_r} = 0.94, \qquad \frac{C_r}{Z_r} \kappa = 6.61
\end{gathered}
\label{eq:constants}
\end{equation}
Both the effective drift $b(\phi)$ and diffusion term $a(\phi)$  are shown in
Fig.~\ref{fig:comparison_lemon} as a reference, as black lines. 

A stochastic trajectory was generated by simulating a diffusion process in the
two-dimensional potential Eq.~(\ref{eq:lemon_slice}), and the binned averages
of Eqs.~(\ref{eq:simple_increm}-\ref{eq:squared_increm}) were computed from the
simulation data, see appendix~\ref{sec:binning}. The time increment $s$ was
chosen to be equal to the integration step. A dictionary $\Theta^{\prime
\prime}$ of $K = 20$ basis functions was
used, with composition given in the Appendix, Eq.~(\ref{eq:theta_2d}).
The dictionary includes the analytic functions for both $b$ and
$a$, $[1, \sin 7\phi]$, Eq.~(\ref{eq:eff_dyn}), to which we refer as analytic
basis functions, as in the previous sections.

The SSR algorithm with cross validation was used to obtain a sparse expression
for the estimated drift and diffusion terms. The results are summarized in
Fig.~\ref{fig:comparison_lemon}, where the same notation as in
Fig.~\ref{fig:degeneracy_1d} is used.

The cross validation score plots in Fig.~\ref{fig:comparison_lemon}a and
Fig.~\ref{fig:comparison_lemon}d allow to locate the $\delta$ transition points
and identify the optimal solutions to Eq.~(\ref{eq:regression_drift}) and
Eq.~(\ref{eq:regression_diffusion}).

As $\delta[n]$ reaches its minimum at solution size $n=1$ for both the drift
and the diffusion, looking at the $\delta[n-1]/\delta[n]$ ratio is  hereby
unnecessary. 

The optimal sparsity values $\tilde{n}_b = 1$ and $\tilde{n}_a = 1$ are
indicated by vertical red lines. The optimal solutions read as:
\begin{align}
\tilde{b}(\phi) = \tilde{c}_b \cdot \Theta(\phi) = 6.39 \sin (7\phi) , \qquad \delta = 4.0 \cdot 10^{-2}\\
\tilde{a}(\phi) = \tilde{c}_a \cdot \Theta(\phi) = 0.95, \qquad \delta = 2.0 \cdot 10^{-4}
\label{eq:b_and_a_optimal}
\end{align}
as shown in the sparsity progress matrices in
Fig.~\ref{fig:comparison_lemon}b and Fig.~\ref{fig:comparison_lemon}e.
 
 \begin{figure}[!ht]
\centering{}
\includegraphics[scale=0.45]{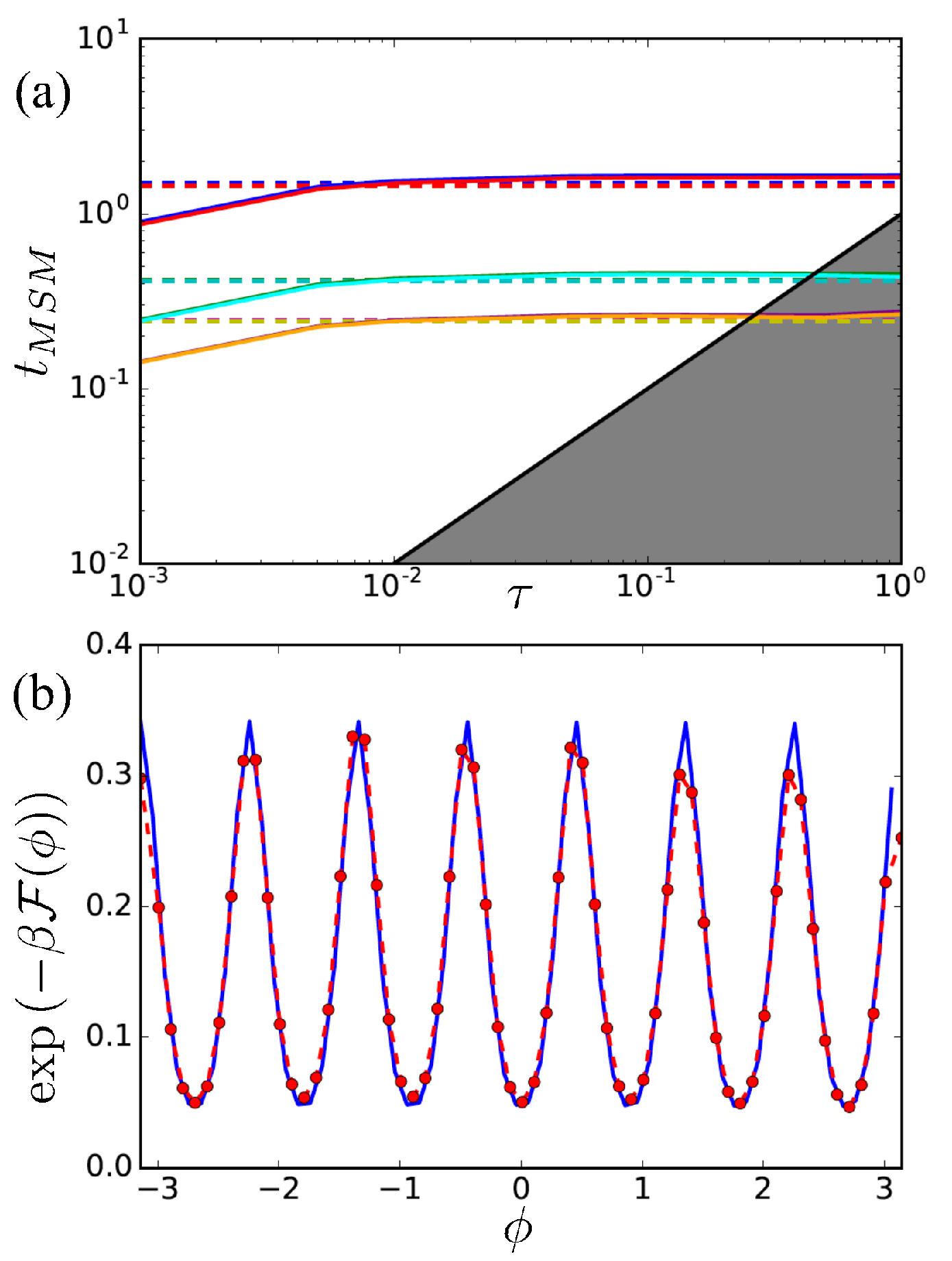}
\caption{Comparison between the kinetics and thermodynamics from both the
analytic and the data-learned $\phi$ projected diffusion. (Top panel) MSM
implied timescales plotted as function of the lag time $\tau$: horizontal dashed
lines indicate the actual timescales of the full system. Deviations
between the two dynamics (solid lines) are minimal. 
(Bottom panel) The stationary distributions $\exp \left(
-\beta\mathcal{F}(\phi)\right) $ of both the analytic (blue solid line) and the reconstructed
dynamics $(\tilde{b}, \tilde{a})$ (dashed line with markers indicating the bin
centers) are compared: oscillation frequency is matched exactly, and there is
a mimimal deviation in the amplitudes because of sampling errors, particularly
close to the top of the oscillation. 
\label{fig:lemon_MSM}
}
\end{figure}

Enforcing sparsity isolates the analytic functions in the dictionary and
makes the solutions of Eq.~(\ref{eq:b_and_a_optimal}) analogous to
Eq.~(\ref{eq:eff_dyn}), as shown in
the comparison plots Fig.~\ref{fig:comparison_lemon}c and
Fig.~\ref{fig:comparison_lemon}f.  

We simulated the ``learned'' projected dynamics $d\phi_t =
\tilde{b}(\phi) dt + \sqrt{2\tilde{a}(\phi) \beta^{-1}}dW_t$ using the same simulation 
parameters as in the original dynamics (Eq.~(\ref{eq:overdamped_langevin}) 
with the potential of Eq.~(\ref{eq:lemon_slice})). We discretized the trajectory data
and performed a Markov State Model analysis~\cite{Prinz2011}.
Fig.~\ref{fig:lemon_MSM}a shows that the first few timescales of the original
dynamics (dashed lines) are
accurately recovered by the learned projected dynamics (solid lines).   Also,
the $\phi$-projected equilibrium distributions of both dynamics agree, as shown
in Fig. ~\ref{fig:lemon_MSM}b.
Thus, the learned projected dynamics recovers both the full thermodynamics and
long timescale kinetics of the original dynamics.

\section{Discussion}
The CV-based sparse regression method SSR presented above appears to be
effective at learning stochastic dynamical equations.
Indeed, the CV analysis allows to identify the optimal solution as
sparsity is maximized while the model predictive power is preserved. The
key features of the drift and the diffusion components (e.g., the dynamics) are
shown to be preserved. 
 
For the one dimensional potential Eq.~(\ref{eq:pot1}), the optimal 
solution closely approximates the gradient of the potential, as shown in Fig.
\ref{fig:comparison_1d}d and Fig. \ref{fig:degeneracy_1d}d.
Similarly, the optimal sparse solution for the lemon slice projected dynamics
reproduces the thermodynamics and the long timescale kinetics of the original model, 
(see Figs.~\ref{fig:comparison_lemon}c,~\ref{fig:comparison_lemon}f,
and~\ref{fig:lemon_MSM}). 

However, the performance of the SSR algorithm and the specific form of the
optimally sparse solution depend on the composition of the dictionary of basis
functions used (e.g., compare $\tilde{\mathbf{c}}$ in Fig.~\ref{fig:comparison_1d}b with 
 $\tilde{\mathbf{c}}^\prime$ in Fig.~\ref{fig:degeneracy_1d}b). 
The ``correct'' solution $\tilde{\theta} = [1, x, x^2, x^3]$ can
be obtained if the solution space is searched greedily, but the SSR algorithm may 
return solutions corresponding to local minima and multiple function dictionaries need
to be considered.

Using ill-conditioned dictionaries containing collinear basis functions 
exacerbates the convergence problem as multiple linear combinations are almost
equivalent, and picking one over others is driven by small perturbations due to
numerical noise. 

Building a dictionary of strictly linearly independent basis functions (such as
Hermite polynomials or Fourier series) is a well established strategy to avoid
the ``many-solution" problem from the very beginning. Another popular approach 
is to run singular value decomposition (SVD) on the regressor matrix
and discard those singular values which are within machine precision (and
therefore contribute exclusively to noise). The reduced set of truncated
singular values naturally defines a space into which the regressor matrix can be
projected. The regression can then be formulated in terms of new effective
variables that are linear combinations of the original ones and are less noisy
by construction; hence, more numerical stability is guaranteed. 
However, if the goal is learning a sparse representation of the potential
energy driving a dynamical system, dictionary entries bear a physical meaning,
and such interpretation may be lost upon SVD or orthogonalization.
For instance, if the input coordinates are composed by a set of contacts or
coordinates or angles in a macromolecule, linear combinations of such
quantities may be far from being physically interpretable. 

The main problem in the convergence of the SSR algorithm is that functions are
pruned from the dictionary at every iteration in a  deterministic fashion:
whenever a given entry is removed from the database, it can not ``resuscitate''
in the next iterations, and the solution may be funneled into a local minimum.
We believe that introducing stochasticity in the pruning of entries and
allowing the reintroduction of previously eliminated entries, resembling Monte
Carlo techniques, could significantly improve the performance of the algorithm and
also compensate (at least partially) for collinearity. 

\begin{acknowledgments}
We are indebted to Frank No\'{e}, Ralf Banisch, Stefan Klus, P\'eter Koltai, and
Steve Brunton for fruitful discussions. 
This work was supported by the National Science Foundation (CHE-1265929,
CHE-1738990, and PHY-1427654), and the Welch Foundation (C-1570). F.N. is a
postdoctoral researcher in the Rice University Academy of Fellows.  Simulations
have been performed on the computer clusters of the Center for Research
Computing at Rice University, supported in part by the Big-Data Private-Cloud
Research Cyberinfrastructure MRI-award (NSF grant CNS-1338099).  
\end{acknowledgments}

\appendix
\section{Details}

\subsection{Binning}
\label{sec:binning}

Solving sparse regression Eq.~(\ref{eq:L0_norm}) usually involves computing,
storing, and inverting large matrices (e.g. $\mathbb{X}$, $\mathbb{Y}$) which
scale linearly with the number of frames in a trajectory $N \gg 1$. If the
dimension of the system is relatively small,  e.g., $d \approx 1$, the problem can be made
more tractable numerically. Let us start by histogramming the coordinate $X$ into
$Q$ bins, i.e.:
\begin{equation}
\left\{X(t_l) \right\}_{l=1, \cdots , N} \mapsto \left\{  \bar{x}_i, w_i  \right\}_{i=1, \cdots , Q}
\label{eq:binning}
\end{equation}
where $\bar{x}_i$  indicates the $i$th bin center and $w_i$ indicates the fraction of data in the $i$th bin,
 which we call \emph{bin weight}.
Subsequently:
$$\mathbb{X}\in\mathbb{R}^{N\times K} \mapsto \mathbb{X}_Q\in\mathbb{R}^{Q\times K}$$
$$\mathbb{Y}\in\mathbb{R}^{N} \mapsto \mathbb{Y}_Q \in\mathbb{R}^Q$$
where $Y_Q$ entries are averaged over each bin.

The sparse regression Eq.~(\ref{eq:L0_norm}) can be cast in the following \emph{weighted} regression
$$\tilde{\mathbf{c}}=\min_{\mathbf{c}} \|\mathbb{W}_Q \mathbb{Y}_Q-\mathbb{W}_Q \mathbb{X}_Q\mathbf{c} \|^{2}_2+\lambda \| \mathbf{c}\|_{0}$$
where the \emph{weight matrix} $\mathbb{W}_Q$ is defined as $$\mathbb{X}_Q = diag \left(w_1, \cdots, w_Q   \right)$$

Both the double well potential and projected dynamics study cases discussed
here are $d=1$ problems, and the binning is used in the results presented in the manuscript.

\subsection{Double well potential}
The data set $\mathcal{D}$ used in the double well potential example consists
of five long independent trajectories of
$N =10^7$ steps each,  generated by integrating the dynamics
Eq.~(\ref{eq:overdamped_langevin}) using a time increment $s =5 \cdot 10^{-3}$
with $m=1$, $k_BT = 1$, $\gamma=1$ (arbitrary units). The simulations were
sufficiently long to ensure sampling from the equilibrium distribution $\pi(x)
\propto \exp(-U(x))$, see Eq.~(\ref{eq:pot1}).

Both the time sequence $\left\{  X(t_l)\right\}_{l=1, \cdots N+1}$ and the set of
simple increments, Eq.~(\ref{eq:simple_increm}): $$\left\{Y_l \right\}_{l
= 1, \cdots , N} = \left\{ \frac{X(t_{l+1}) - X(t_l)}{s} \right\}_{l = 1, \cdots , N}$$  were discretized into
$Q = 90$ bins along the $x$ axis, giving rise to an increment matrix $\mathbb{Y} \in \mathbb{R}^Q$

Two different $K = 20$ basis function dictionaries $\Theta$ and $\Theta^\prime$ were
considered. Both contain the four functions $[1, x, x^2, x^3]$ entering Eq.~(\ref{eq:pot1}), the
remaining entries were selected randomly from a larger set of $100$ basis
functions.  The specific composition of the two dictionaries read:
\begin{widetext}
\begin{centering}
\begin{align}
\begin{split}
\Theta(x) =  & [    1,  \quad x,  \quad x^2,  \quad x^3,   \quad x^4,  \quad
x^5,  \quad, x^6,  \quad x^7,  \quad x^8,  \quad x^9,   \quad x^{10},  \quad
\sin x, \quad \cos x, \\ 
& \sin (6x), \quad \cos (6x), \quad \sin (11x), \quad \cos (11x), \quad \tanh
(10x), \quad -10 \tanh^2(10x) + 10 e^{-50x^2} ]
\label{eq:theta}
\end{split}
\end{align}

\begin{align}
\begin{split}
\Theta^\prime(x) = &  [ 1,  \quad x,  \quad x^2, \quad x^3, \quad \sin x, 
\quad \cos 11x,  \quad \sin 11x,  \quad -10 \tanh^2(10x) + 10,  \\ 
&  -10 \tanh^2(10x-10) + 10, \quad e^{-50x^2},  \quad e^{-50(x-3)^2}, \\ 
& e^{-0.3x^2}, \quad e^{-0.3(x-3)^2}, \quad e^{-2(x-2)^2}, \quad e^{-2(x-4)^2},
\quad e^{-50(x-4)^2}, \\ &  e^{-0.6(x-4)^2}, \quad e^{-0.6(x-3)^2},\quad -2
\tanh^2(2x-4) + 2, \quad  \tanh^2(x-4) + 1 ]
\label{eq:thetaprime}
\end{split}
\end{align}
\end{centering}
\end{widetext}
Each dictionary computed on the binned coordinate generates a database
$\mathbb{X} \in \mathbb{R}^{Q \times K}: X_{ij} = \Theta_i(\bar{x}_j)$, $\bar{x}_j$ being the value of the
coordinate in the $j$-th bin, Eq.~(\ref{eq:binning}). 

$N_k = 50$ independent cross validation calculations were run and the cross validation scores averaged:
$$\delta^2 = \frac{1}{N_k}\sum_{k=1}^{N_k} \delta_k^2$$
where $\delta_k$ is given by Eq.~(\ref{eq:CV_score}). Each CV run, $i$,
is associated with its own decomposition of the data set into folds $$\mathcal{D}
\rightarrow \left\{A^j_i \right\}_{j=1, \cdots, 5}.$$

\subsection{Greedy sparsity search}
We provide a short description of the parameters used in the greedy search in
the solution space for the double well potential. 

The number of independent dictionary combinations reads
\[
 m_n = 
  \begin{cases} 
   C_{M,n} & \text{if } n \in [2,3,4] \\
   10^5       & \text{otherwise}
  \end{cases}
\]
Cross validation scores were computed by running $N_k = 20$ independent $5$-fold cross validations, and 
averaging over all runs as already mentioned.

\subsection{Projected dynamics}

The dataset $\mathcal{D}$ for the lemon slice example consists of one single 
$N = 10^7$ step diffusive
trajectory, generated by integrating dynamics Eq.~(\ref{eq:overdamped_langevin})
for $U = U(x,y)$ using a time increment $s = 10^{-3}$ with $m=1$, $k_BT = 1$,
$\gamma=1$ (arbitrary units). The simulation was long enough to ensure
equilibrium sampling. The projection coordinate is the polar angle $\xi = \phi$.

Both the time sequence $\left\{ \phi(x(t_l), y(t_l))  \right\}_{l=1, \cdots ,
N+1}$, the simple increments Eq.~(\ref{eq:simple_increm}) $$\left\{ Y_b(\phi) \right\}_{l=1, \cdots ,
N} = \left\{\frac{\phi(t_{l+1}) - \phi(t_l)}{s} \right\}_{l=1, \cdots ,N}$$  and the squared
increments  Eq.~(\ref{eq:squared_increm}) $$\left\{ Y_a(\phi) \right\}_{l=1, \cdots ,
N} = \left\{ \frac{1}{2} \frac{(\phi(t_{l+1}) - \phi(t_l))^2}{s} \right\}_{l = 1, \cdots, N}$$  were
discretized into $63$ bins along the $\phi \in [-\pi, \pi]$ axis, giving rise to $\mathbb{Y}_a,\ \mathbb{Y}_b \in \mathbb{R}^Q$.

A $K = 20$ basis function dictionary $\Theta^{\prime \prime}$ was considered,
such that $[1, \sin 7x] \in \Theta^{\prime \prime}$,
Eq.~(\ref{eq:eff_dyn}). Its specific composition reads:
\begin{widetext}
\begin{centering}
\begin{align}
\begin{split}
\Theta^{\prime \prime}(x) =  & [  1,  \quad x,  \quad x^2,  \quad x^3,  \quad x^4,  \quad x^5,  \quad, x^6,  \quad \sin x, \quad \cos x \\ 
& \sin 4x, \quad \cos 4x, \quad \sin 7x, \quad \cos (7x), \quad \tanh (10x), \\ 
&  -10 \tanh^2(10x) + 10, \quad e^{-50x^2}, \quad \tanh(2x), \quad -2\tanh^2(2x)+2, \\ 
& e^{-2x^2}, \quad \tanh(x) ]
\label{eq:theta_2d}
\end{split}
\end{align}
\end{centering}
\end{widetext}
The functions in the dictionary computed on the binned coordinate generate a database
$\mathbb{X} \in \mathbb{R}^{Q \times K}: X_{ij} = \Theta^{\prime \prime}_i(\bar{\phi}_j)$, $\bar{\phi}_j$ being the value
of the coordinate in the $j$-th angular bin (notation from Eq.~(\ref{eq:binning})). 

Cross validation scores were computed by running $N_k = 50$ independent
$7$-fold cross validations, and averaging over all runs as already mentioned.

\section{Proof of Theorem 1}

\label{sec:proof_thm1}

In order to prove Theorem 1, we need the following non-linear version of
Fubini's theorem, called the co-area formula:

\begin{eqnarray*}
\int_{\mathbb{R}^{n}}f(x)\,\mathrm{dx} & = & \int_{\mathbb{R}^{m}}\int_{\Sigma_{z}}f(x)J^{-1/2}(x)\,\mathrm{d}\sigma_{z}(x)\,\mathrm{d}z.
\end{eqnarray*}
The co-area formula implies that for functions $f,\,g$ which only
depend on $z$, we have

\begin{eqnarray}
\int_{\mathbb{R}^{n}}f(x)g(x)\mu(x)\,\mathrm{d}x & = & \int_{\mathbb{R}^{m}}f(z)g(z)\nu(z)\,\mathrm{d}z.\label{eq:duality_scalar_products}
\end{eqnarray}
Now, regression problems Eqs.~(\ref{eq:regression_drift}-\ref{eq:regression_diffusion})
are equivalent to the normal equations

\begin{eqnarray}
\Theta^{T}\Theta c_{i} & = & \Theta^{T}Y_{i},\label{eq:normal_eqs_drift}\\
\Theta^{T}\Theta c_{ij} & = & \Theta^{T}Y_{ij}.\label{eq:normal_eqs_diff}
\end{eqnarray}
By ergodicity of the process and Eq.~(\ref{eq:duality_scalar_products}),

\begin{eqnarray*}
\frac{1}{L}\left[\Theta^{T}\Theta\right]_{k,k'} & = & \frac{1}{L}\sum_{l=1}^{L}f_{k}(X(t_{l}))f_{k'}(X(t_{l}))\\
 & \xrightarrow{L\rightarrow\infty} & \langle f_{k},f_{k'}\rangle_{\mu}\\
 & = & \langle f_{k},f_{k'}\rangle_{\nu}.
\end{eqnarray*}
To calculte the asymptotic limits of the right-hand sides in
Eqs.~(\ref{eq:normal_eqs_drift}-\ref{eq:normal_eqs_diff}), we use the
conditional transition probability density $p^{s}(x,y)$ over time
$s$, and introduce the quantities

\begin{eqnarray}
d_{i}^{s}(x) & = & \int_{\mathbb{R}^{n}}\frac{1}{s}\left[\xi_{i}(y)-\xi_{i}(x)\right]p^{s}(x,y)\,\mathrm{d}y,\label{eq:definition_di}\\
d_{ij}^{s}(x) & = & \int_{\mathbb{R}^{n}}\frac{1}{s}\left[\xi_{i}(y)-\xi_{i}(x)\right]\left[\xi_{j}(y)-\xi_{j}(x)\right]p^{s}(x,y)\,\mathrm{d}y.\label{eq:definition_dij}
\end{eqnarray}
We find for the right-hand side of Eq.~(\ref{eq:normal_eqs_drift}):

\begin{widetext}
\begin{eqnarray*}
\frac{1}{L}\left[\Theta^{T}Y_{i}\right]_{k} & = & \frac{1}{L}\sum_{l=1}^{L}f_{l}(X(t_{l}))\frac{1}{s}\left[\xi_{i}(X(t_{l+1})-\xi_{i}(X(t_{l}))\right]\\
 & \xrightarrow{L\rightarrow\infty} & \int_{\mathbb{R}^{n}}\int_{\mathbb{R}^{n}}\frac{1}{s}\left[\xi_{i}(y)-\xi_{i}(x)\right]p^{s}(x,y)\mu(x)f_{k}(x)\,\mathrm{d}x\,\mathrm{d}y\\
 & = & \int_{\mathbb{R}^{n}}d_{i}^{s}(x)f_{k}(x)\mu(x)\,\mathrm{d}x\\
 & = & \int_{\mathbb{R}^{m}}\left[\int_{\Sigma_{z}}d_{i}^{s}(x)\mu(x)J^{-1/2}(x)\,\mathrm{d}\sigma_{z}(x)\right]f_{k}(z)\,\mathrm{d}z\\
 & = & \int_{\mathbb{R}^{m}}\left[\int_{\Sigma_{z}}d_{i}^{s}(x)\,\mathrm{d}\mu_{z}(x)\right]f_{k}(z)\nu(z)\,\mathrm{d}z\\
 & \xrightarrow{s\rightarrow0} & \langle b_{i}^{\xi},f_{k}\rangle_{\nu}.
\end{eqnarray*}
 \end{widetext}
In the fourth line, we have used the co-area formula, followed by
the definition of the restricted equilibrium measure $\mu_{z}$ in
Eq.~(\ref{eq:restricted_eq_measure}). The last step follows from
the Kramers-Moyal formula Eq.~(\ref{eq:Kramers_Moyal_eff_1}). In
the same way, we find that

\begin{eqnarray*}
\frac{1}{L}\left[\Theta^{T}Y_{ij}\right]_{k} & \xrightarrow{L\rightarrow\infty} & \int_{\mathbb{R}^{n}}d_{ij}^{s}(x)f_{k}(x)\mu(x)\,\mathrm{d}x\\
 & \xrightarrow{s\rightarrow0} & \langle a_{ij}^{\xi},f_{k}\rangle_{\nu}.
\end{eqnarray*}
We conclude that Eqs.~(\ref{eq:normal_eqs_drift}-\ref{eq:normal_eqs_diff})
converge to the $L_{\nu}^{2}$-best-approximation problem for the
effective drift and diffusion, \textbf{$b_{i}^{\xi}$} and $a_{ij}^{\xi}$,
using the basis set $f_{k}$.

\section{Estimation of Potential Energy}

\label{sec:energy_relation}

Here, we show that the regression problem Eq.~(\ref{eq:regression_energy})
converges to the best-approximation problem for the generalized energy
Eq.~(\ref{eq:drift_diffusion_energy}) from the basis set of vector
fields $\nabla f_{k},\,k=1,\ldots,K$. The case where no projection
is applied can be recovered by choosing $\xi$ as the identity on
$\mathbb{R}^{n}$, and the formulation we introduced for the overdamped
Langevin dynamics is obtained by setting the diffusion to an identity
matrix.

The proof is very similar to the previous section. Starting from the
normal equation

\begin{eqnarray*}
D^{T}Dv & = & D^{T}Y,
\end{eqnarray*}
we first find that

\begin{eqnarray*}
\frac{1}{L}\left[D^{T}D\right]_{k,k'} & = & \frac{1}{L}\sum_{i,l}\frac{\partial f_{k}}{\partial z_{i}}(X(t_{l}))\frac{\partial f_{k'}}{\partial z_{i}}(X(t_{l}))\\
 & \xrightarrow{L\rightarrow\infty} & \sum_{i=1}^{m}\langle\frac{\partial f_{k}}{\partial z_{i}},\frac{\partial f_{k'}}{\partial z_{i}}\rangle_{\mu}\\
 & = & \langle\nabla f_{k},\nabla f_{k'}\rangle_{\nu}.
\end{eqnarray*}
Moreover, the data vector $D^{T}Y$ converges to

\begin{widetext}
\begin{eqnarray*}
\frac{1}{L}\left[D^{T}Y\right]_{k} & = & \frac{1}{L}\sum_{i,l}\frac{\partial f_{k}}{\partial z_{i}}(X(t_{l}))\left[\left(a^{\xi}\right)^{-1}(X(t_{l}))(\frac{1}{\beta}\nabla\cdot a_{i}^{\xi}(X(t_{l})) - e^{s}(X(t_{l+1}), X(t_{l})))\right]_{i}\\
 & \xrightarrow{L\rightarrow\infty} & \sum_{i=1}^{m}\int\int\frac{\partial f_{k}}{\partial z_{i}}(x)\left[\left(a^{\xi}\right)^{-1}(x)(\frac{1}{\beta}\nabla\cdot a_{i}^{\xi}(x) - e^{s}(y,x))\right]_{i}\mu(x)p^{s}(x,y)\,\mathrm{d}x\,\mathrm{d}y.
\end{eqnarray*}
\end{widetext}
We can proceed exactly as in the previous section, recalling the vectors
$d^{s}$ from Eq.~(\ref{eq:definition_di}) and the relation between
drift and diffusion, Eq.~(\ref{eq:drift_diffusion_energy}):

\begin{widetext}
\begin{eqnarray*}
\frac{1}{L}\left[D^{T}Y\right]_{k} & = & \sum_{i=1}^{m}\int\frac{\partial f_{k}}{\partial z_{i}}(x)\left[\left(a^{\xi}\right)^{-1}(x)(\frac{1}{\beta}\nabla\cdot a_{i}^{\xi}(x) - d^{s}(x))\right]_{i}\mu(x)\,\mathrm{d}x\\
 & = & \sum_{i=1}^{m}\int_{\mathbb{R}^{m}}\frac{\partial f_{k}}{\partial z_{i}}(z)\left[\left(a^{\xi}\right)^{-1}(z)(\frac{1}{\beta}\nabla\cdot a_{i}^{\xi}(z) - \int_{\Sigma_{z}}d^{s}(x)\,\mathrm{d}\mu_{z}(x))\right]_{i}\nu(z)\,\mathrm{d}z\\
 & \xrightarrow{s\rightarrow0} & \sum_{i=1}^{M}\langle\frac{\partial f_{k}}{\partial z_{i}},\left[\left(a^{\xi}\right)^{-1}(\frac{1}{\beta}\nabla\cdot a_{i}^{\xi} - b^{\xi})\right]_{i}\rangle_{\nu}\\
 & = & \langle\nabla f_{k},\nabla \mathcal{F}\rangle_{\nu}.
\end{eqnarray*}
\end{widetext}


\begin{thebibliography}{23}%
\makeatletter
\providecommand \@ifxundefined [1]{%
 \@ifx{#1\undefined}
}%
\providecommand \@ifnum [1]{%
 \ifnum #1\expandafter \@firstoftwo
 \else \expandafter \@secondoftwo
 \fi
}%
\providecommand \@ifx [1]{%
 \ifx #1\expandafter \@firstoftwo
 \else \expandafter \@secondoftwo
 \fi
}%
\providecommand \natexlab [1]{#1}%
\providecommand \enquote  [1]{``#1''}%
\providecommand \bibnamefont  [1]{#1}%
\providecommand \bibfnamefont [1]{#1}%
\providecommand \citenamefont [1]{#1}%
\providecommand \href@noop [0]{\@secondoftwo}%
\providecommand \href [0]{\begingroup \@sanitize@url \@href}%
\providecommand \@href[1]{\@@startlink{#1}\@@href}%
\providecommand \@@href[1]{\endgroup#1\@@endlink}%
\providecommand \@sanitize@url [0]{\catcode `\\12\catcode `\$12\catcode
  `\&12\catcode `\#12\catcode `\^12\catcode `\_12\catcode `\%12\relax}%
\providecommand \@@startlink[1]{}%
\providecommand \@@endlink[0]{}%
\providecommand \url  [0]{\begingroup\@sanitize@url \@url }%
\providecommand \@url [1]{\endgroup\@href {#1}{\urlprefix }}%
\providecommand \urlprefix  [0]{URL }%
\providecommand \Eprint [0]{\href }%
\providecommand \doibase [0]{http://dx.doi.org/}%
\providecommand \selectlanguage [0]{\@gobble}%
\providecommand \bibinfo  [0]{\@secondoftwo}%
\providecommand \bibfield  [0]{\@secondoftwo}%
\providecommand \translation [1]{[#1]}%
\providecommand \BibitemOpen [0]{}%
\providecommand \bibitemStop [0]{}%
\providecommand \bibitemNoStop [0]{.\EOS\space}%
\providecommand \EOS [0]{\spacefactor3000\relax}%
\providecommand \BibitemShut  [1]{\csname bibitem#1\endcsname}%
\let\auto@bib@innerbib\@empty
\bibitem [{\citenamefont {No\'e}\ and\ \citenamefont
  {Clementi}(2017)}]{NoeClementiCOSB2017}%
  \BibitemOpen
  \bibfield  {author} {\bibinfo {author} {\bibfnamefont {F.}~\bibnamefont
  {No\'e}}\ and\ \bibinfo {author} {\bibfnamefont {C.}~\bibnamefont
  {Clementi}},\ }\href@noop {} {\bibfield  {journal} {\bibinfo  {journal}
  {Curr. Opin. Struct. Biol.}\ }\textbf {\bibinfo {volume} {43}},\ \bibinfo
  {pages} {141} (\bibinfo {year} {2017})}\BibitemShut {NoStop}%
\bibitem [{\citenamefont {Rohrdanz}, \citenamefont {Zheng},\ and\ \citenamefont
  {Clementi}(2013)}]{RohrdanzEtAl_AnnRevPhysChem13_MountainPasses}%
  \BibitemOpen
  \bibfield  {author} {\bibinfo {author} {\bibfnamefont {M.~A.}\ \bibnamefont
  {Rohrdanz}}, \bibinfo {author} {\bibfnamefont {W.}~\bibnamefont {Zheng}}, \
  and\ \bibinfo {author} {\bibfnamefont {C.}~\bibnamefont {Clementi}},\
  }\href@noop {} {\bibfield  {journal} {\bibinfo  {journal} {Ann. Rev. Phys.
  Chem.}\ }\textbf {\bibinfo {volume} {64}},\ \bibinfo {pages} {295} (\bibinfo
  {year} {2013})}\BibitemShut {NoStop}%
\bibitem [{\citenamefont {Kevrekidis}\ and\ \citenamefont
  {Samaey}(2009)}]{Kevrekidis2009}%
  \BibitemOpen
  \bibfield  {author} {\bibinfo {author} {\bibfnamefont {I.~G.}\ \bibnamefont
  {Kevrekidis}}\ and\ \bibinfo {author} {\bibfnamefont {G.}~\bibnamefont
  {Samaey}},\ }\href {\doibase 10.1146/annurev.physchem.59.032607.093610}
  {\bibfield  {journal} {\bibinfo  {journal} {Ann. Rev. Phys. Chem.}\ }\textbf
  {\bibinfo {volume} {60}},\ \bibinfo {pages} {321} (\bibinfo {year}
  {2009})}\BibitemShut {NoStop}%
\bibitem [{\citenamefont {Brunton}, \citenamefont {Proctor},\ and\
  \citenamefont {Kutz}(2016)}]{BruntonPNAS2016}%
  \BibitemOpen
  \bibfield  {author} {\bibinfo {author} {\bibfnamefont {S.~L.}\ \bibnamefont
  {Brunton}}, \bibinfo {author} {\bibfnamefont {J.~L.}\ \bibnamefont
  {Proctor}}, \ and\ \bibinfo {author} {\bibfnamefont {J.~N.}\ \bibnamefont
  {Kutz}},\ }\href {\doibase 10.1073/pnas.1517384113} {\bibfield  {journal}
  {\bibinfo  {journal} {Proc. Natl. Acad. Sci. USA}\ }\textbf {\bibinfo
  {volume} {113}},\ \bibinfo {pages} {3932} (\bibinfo {year}
  {2016})}\BibitemShut {NoStop}%
\bibitem [{\citenamefont {Tibshirani}(1996)}]{Tibshirani96Lasso}%
  \BibitemOpen
  \bibfield  {author} {\bibinfo {author} {\bibfnamefont {R.}~\bibnamefont
  {Tibshirani}},\ }\href@noop {} {\bibfield  {journal} {\bibinfo  {journal} {J.
  R. Stat. Soc. B}\ }\textbf {\bibinfo {volume} {58}},\ \bibinfo {pages} {267}
  (\bibinfo {year} {1996})}\BibitemShut {NoStop}%
\bibitem [{\citenamefont {James}\ \emph {et~al.}(2013)\citenamefont {James},
  \citenamefont {Witten}, \citenamefont {Hastie},\ and\ \citenamefont
  {Tibshirani}}]{James2013}%
  \BibitemOpen
  \bibfield  {author} {\bibinfo {author} {\bibfnamefont {G.}~\bibnamefont
  {James}}, \bibinfo {author} {\bibfnamefont {D.}~\bibnamefont {Witten}},
  \bibinfo {author} {\bibfnamefont {T.}~\bibnamefont {Hastie}}, \ and\ \bibinfo
  {author} {\bibfnamefont {R.}~\bibnamefont {Tibshirani}},\ }\href@noop {}
  {\emph {\bibinfo {title} {An Introduction to Statistical Learning}}}\
  (\bibinfo  {publisher} {Springer New York},\ \bibinfo {year}
  {2013})\BibitemShut {NoStop}%
\bibitem [{\citenamefont {Donoho}(2006)}]{Donoho2006}%
  \BibitemOpen
  \bibfield  {author} {\bibinfo {author} {\bibfnamefont {D.}~\bibnamefont
  {Donoho}},\ }\href@noop {} {\bibfield  {journal} {\bibinfo  {journal} {{IEEE}
  Trans. Inf. Theory}\ }\textbf {\bibinfo {volume} {52}},\ \bibinfo {pages}
  {1289} (\bibinfo {year} {2006})}\BibitemShut {NoStop}%
\bibitem [{\citenamefont {Baraniuk}(2007)}]{Baraniuk2007}%
  \BibitemOpen
  \bibfield  {author} {\bibinfo {author} {\bibfnamefont {R.}~\bibnamefont
  {Baraniuk}},\ }\href@noop {} {\bibfield  {journal} {\bibinfo  {journal}
  {{IEEE} Signal Process. Mag.}\ }\textbf {\bibinfo {volume} {24}},\ \bibinfo
  {pages} {118} (\bibinfo {year} {2007})}\BibitemShut {NoStop}%
\bibitem [{\citenamefont {Rudy}\ \emph {et~al.}(2017)\citenamefont {Rudy},
  \citenamefont {Brunton}, \citenamefont {Proctor},\ and\ \citenamefont
  {Kutz}}]{Rudye1602614}%
  \BibitemOpen
  \bibfield  {author} {\bibinfo {author} {\bibfnamefont {S.~H.}\ \bibnamefont
  {Rudy}}, \bibinfo {author} {\bibfnamefont {S.~L.}\ \bibnamefont {Brunton}},
  \bibinfo {author} {\bibfnamefont {J.~L.}\ \bibnamefont {Proctor}}, \ and\
  \bibinfo {author} {\bibfnamefont {J.~N.}\ \bibnamefont {Kutz}},\ }\href@noop
  {} {\bibfield  {journal} {\bibinfo  {journal} {Sci. Adv.}\ }\textbf {\bibinfo
  {volume} {3}} (\bibinfo {year} {2017})}\BibitemShut {NoStop}%
\bibitem [{\citenamefont {Legoll}\ and\ \citenamefont
  {Lelièvre}(2010)}]{Legoll_Nonlinearity2010}%
  \BibitemOpen
  \bibfield  {author} {\bibinfo {author} {\bibfnamefont {F.}~\bibnamefont
  {Legoll}}\ and\ \bibinfo {author} {\bibfnamefont {T.}~\bibnamefont
  {Lelièvre}},\ }\href {http://stacks.iop.org/0951-7715/23/i=9/a=006}
  {\bibfield  {journal} {\bibinfo  {journal} {Nonlinearity}\ }\textbf {\bibinfo
  {volume} {23}},\ \bibinfo {pages} {2131} (\bibinfo {year}
  {2010})}\BibitemShut {NoStop}%
\bibitem [{\citenamefont {Zhang}, \citenamefont {Hartmann},\ and\ \citenamefont
  {Sch\"{u}tte}(2016)}]{Zhang_Faraday16}%
  \BibitemOpen
  \bibfield  {author} {\bibinfo {author} {\bibfnamefont {W.}~\bibnamefont
  {Zhang}}, \bibinfo {author} {\bibfnamefont {C.}~\bibnamefont {Hartmann}}, \
  and\ \bibinfo {author} {\bibfnamefont {C.}~\bibnamefont {Sch\"{u}tte}},\
  }\href@noop {} {\bibfield  {journal} {\bibinfo  {journal} {Faraday Discuss.}\
  }\textbf {\bibinfo {volume} {195}},\ \bibinfo {pages} {365} (\bibinfo {year}
  {2016})}\BibitemShut {NoStop}%
\bibitem [{\citenamefont {Risken}\ and\ \citenamefont
  {Haken}(1989)}]{Risken1989}%
  \BibitemOpen
  \bibfield  {author} {\bibinfo {author} {\bibfnamefont {H.}~\bibnamefont
  {Risken}}\ and\ \bibinfo {author} {\bibfnamefont {H.}~\bibnamefont {Haken}},\
  }\href@noop {} {\emph {\bibinfo {title} {{The Fokker-Planck Equation: Methods
  of Solution and Applications Second Edition}}}}\ (\bibinfo  {publisher}
  {Springer},\ \bibinfo {year} {1989})\BibitemShut {NoStop}%
\bibitem [{\citenamefont {Pavliotis}(2014)}]{Pavliotis:2014aa}%
  \BibitemOpen
  \bibfield  {author} {\bibinfo {author} {\bibfnamefont {G.}~\bibnamefont
  {Pavliotis}},\ }\href@noop {} {\emph {\bibinfo {title} {Stochastic Processes
  and Applications: Diffusion Processes, the Fokker-Planck and Langevin
  Equations}}},\ Texts in Applied Mathematics\ (\bibinfo  {publisher} {Springer
  New York},\ \bibinfo {year} {2014})\BibitemShut {NoStop}%
\bibitem [{\citenamefont {Tibshirani}(2011)}]{Tibshirani_perspective}%
  \BibitemOpen
  \bibfield  {author} {\bibinfo {author} {\bibfnamefont {R.}~\bibnamefont
  {Tibshirani}},\ }\href@noop {} {\bibfield  {journal} {\bibinfo  {journal} {J.
  Royal Stat. Soc. B}\ }\textbf {\bibinfo {volume} {73}},\ \bibinfo {pages}
  {273} (\bibinfo {year} {2011})}\BibitemShut {NoStop}%
\bibitem [{\citenamefont {Mallat}\ and\ \citenamefont
  {Zhang}(1993)}]{matching_pursuit}%
  \BibitemOpen
  \bibfield  {author} {\bibinfo {author} {\bibfnamefont {S.~G.}\ \bibnamefont
  {Mallat}}\ and\ \bibinfo {author} {\bibfnamefont {Z.}~\bibnamefont {Zhang}},\
  }\href@noop {} {\bibfield  {journal} {\bibinfo  {journal} {IEEE Transactions
  on Signal Processing}\ }\textbf {\bibinfo {volume} {41}},\ \bibinfo {pages}
  {3397} (\bibinfo {year} {1993})}\BibitemShut {NoStop}%
\bibitem [{\citenamefont {Pati}, \citenamefont {Rezaiifar},\ and\ \citenamefont
  {Krishnaprasad}(1993)}]{OMP}%
  \BibitemOpen
  \bibfield  {author} {\bibinfo {author} {\bibfnamefont {Y.~C.}\ \bibnamefont
  {Pati}}, \bibinfo {author} {\bibfnamefont {R.}~\bibnamefont {Rezaiifar}}, \
  and\ \bibinfo {author} {\bibfnamefont {P.~S.}\ \bibnamefont
  {Krishnaprasad}},\ }in\ \href@noop {} {\emph {\bibinfo {booktitle}
  {Proceedings of 27th Asilomar Conference on Signals, Systems and
  Computers}}}\ (\bibinfo {year} {1993})\ pp.\ \bibinfo {pages}
  {40--44}\BibitemShut {NoStop}%
\bibitem [{\citenamefont {Zou}\ and\ \citenamefont
  {Hastie}(2005)}]{ElasticNet}%
  \BibitemOpen
  \bibfield  {author} {\bibinfo {author} {\bibfnamefont {H.}~\bibnamefont
  {Zou}}\ and\ \bibinfo {author} {\bibfnamefont {T.}~\bibnamefont {Hastie}},\
  }\href@noop {} {\bibfield  {journal} {\bibinfo  {journal} {J. Royal Stat.
  Soc. B}\ }\textbf {\bibinfo {volume} {67}},\ \bibinfo {pages} {301} (\bibinfo
  {year} {2005})}\BibitemShut {NoStop}%
\bibitem [{\citenamefont {Hastie}, \citenamefont {Tibshirani},\ and\
  \citenamefont {Friedman}(2001)}]{hastie_book}%
  \BibitemOpen
  \bibfield  {author} {\bibinfo {author} {\bibfnamefont {T.}~\bibnamefont
  {Hastie}}, \bibinfo {author} {\bibfnamefont {R.}~\bibnamefont {Tibshirani}},
  \ and\ \bibinfo {author} {\bibfnamefont {J.}~\bibnamefont {Friedman}},\
  }\href@noop {} {\emph {\bibinfo {title} {The Elements of Statistical
  Learning}}},\ Springer Series in Statistics\ (\bibinfo  {publisher} {Springer
  New York Inc.},\ \bibinfo {address} {New York, NY, USA},\ \bibinfo {year}
  {2001})\BibitemShut {NoStop}%
\bibitem [{\citenamefont {Kohavi}(1995)}]{Kohavi95astudy}%
  \BibitemOpen
  \bibfield  {author} {\bibinfo {author} {\bibfnamefont {R.}~\bibnamefont
  {Kohavi}},\ }in\ \href@noop {} {\emph {\bibinfo {booktitle} {Proceedings of
  the 14th International Joint Conference on Artificial Intelligence - Volume
  2}}},\ \bibinfo {series and number} {IJCAI'95}\ (\bibinfo  {publisher}
  {Morgan Kaufmann Publishers Inc.},\ \bibinfo {address} {San Francisco, CA,
  USA},\ \bibinfo {year} {1995})\ pp.\ \bibinfo {pages}
  {1137--1143}\BibitemShut {NoStop}%
\bibitem [{\citenamefont {Geisser}(1975)}]{geisser1975predictive}%
  \BibitemOpen
  \bibfield  {author} {\bibinfo {author} {\bibfnamefont {S.}~\bibnamefont
  {Geisser}},\ }\href@noop {} {\bibfield  {journal} {\bibinfo  {journal}
  {Journal of the American Statistical Association}\ }\textbf {\bibinfo
  {volume} {70}},\ \bibinfo {pages} {320} (\bibinfo {year} {1975})}\BibitemShut
  {NoStop}%
\bibitem [{\citenamefont {Pedregosa}\ \emph {et~al.}(2011)\citenamefont
  {Pedregosa}, \citenamefont {Varoquaux}, \citenamefont {Gramfort},
  \citenamefont {Michel}, \citenamefont {Thirion}, \citenamefont {Grisel},
  \citenamefont {Blondel}, \citenamefont {Prettenhofer}, \citenamefont {Weiss},
  \citenamefont {Dubourg}, \citenamefont {Vanderplas}, \citenamefont {Passos},
  \citenamefont {Cournapeau}, \citenamefont {Brucher}, \citenamefont {Perrot},\
  and\ \citenamefont {Duchesnay}}]{scikit-learn}%
  \BibitemOpen
  \bibfield  {author} {\bibinfo {author} {\bibfnamefont {F.}~\bibnamefont
  {Pedregosa}}, \bibinfo {author} {\bibfnamefont {G.}~\bibnamefont
  {Varoquaux}}, \bibinfo {author} {\bibfnamefont {A.}~\bibnamefont {Gramfort}},
  \bibinfo {author} {\bibfnamefont {V.}~\bibnamefont {Michel}}, \bibinfo
  {author} {\bibfnamefont {B.}~\bibnamefont {Thirion}}, \bibinfo {author}
  {\bibfnamefont {O.}~\bibnamefont {Grisel}}, \bibinfo {author} {\bibfnamefont
  {M.}~\bibnamefont {Blondel}}, \bibinfo {author} {\bibfnamefont
  {P.}~\bibnamefont {Prettenhofer}}, \bibinfo {author} {\bibfnamefont
  {R.}~\bibnamefont {Weiss}}, \bibinfo {author} {\bibfnamefont
  {V.}~\bibnamefont {Dubourg}}, \bibinfo {author} {\bibfnamefont
  {J.}~\bibnamefont {Vanderplas}}, \bibinfo {author} {\bibfnamefont
  {A.}~\bibnamefont {Passos}}, \bibinfo {author} {\bibfnamefont
  {D.}~\bibnamefont {Cournapeau}}, \bibinfo {author} {\bibfnamefont
  {M.}~\bibnamefont {Brucher}}, \bibinfo {author} {\bibfnamefont
  {M.}~\bibnamefont {Perrot}}, \ and\ \bibinfo {author} {\bibfnamefont
  {E.}~\bibnamefont {Duchesnay}},\ }\href@noop {} {\bibfield  {journal}
  {\bibinfo  {journal} {J. Mach. Learn. Res.}\ }\textbf {\bibinfo {volume}
  {12}},\ \bibinfo {pages} {2825} (\bibinfo {year} {2011})}\BibitemShut
  {NoStop}%
\bibitem [{\citenamefont {Bittracher}\ \emph {et~al.}(2017)\citenamefont
  {Bittracher}, \citenamefont {Koltai}, \citenamefont {Klus}, \citenamefont
  {Banisch}, \citenamefont {Dellnitz},\ and\ \citenamefont
  {Sch{\"u}tte}}]{Bittracher2017}%
  \BibitemOpen
  \bibfield  {author} {\bibinfo {author} {\bibfnamefont {A.}~\bibnamefont
  {Bittracher}}, \bibinfo {author} {\bibfnamefont {P.}~\bibnamefont {Koltai}},
  \bibinfo {author} {\bibfnamefont {S.}~\bibnamefont {Klus}}, \bibinfo {author}
  {\bibfnamefont {R.}~\bibnamefont {Banisch}}, \bibinfo {author} {\bibfnamefont
  {M.}~\bibnamefont {Dellnitz}}, \ and\ \bibinfo {author} {\bibfnamefont
  {C.}~\bibnamefont {Sch{\"u}tte}},\ }\href@noop {} {\bibfield  {journal}
  {\bibinfo  {journal} {J. Nonlinear Science}\ } (\bibinfo {year}
  {2017})}\BibitemShut {NoStop}%
\bibitem [{\citenamefont {Prinz}\ \emph {et~al.}(2011)\citenamefont {Prinz},
  \citenamefont {Wu}, \citenamefont {Sarich}, \citenamefont {Keller},
  \citenamefont {Senne}, \citenamefont {Held}, \citenamefont {Chodera},
  \citenamefont {Sch{\"{u}}tte},\ and\ \citenamefont {No{\'{e}}}}]{Prinz2011}%
  \BibitemOpen
  \bibfield  {author} {\bibinfo {author} {\bibfnamefont {J.-H.}\ \bibnamefont
  {Prinz}}, \bibinfo {author} {\bibfnamefont {H.}~\bibnamefont {Wu}}, \bibinfo
  {author} {\bibfnamefont {M.}~\bibnamefont {Sarich}}, \bibinfo {author}
  {\bibfnamefont {B.}~\bibnamefont {Keller}}, \bibinfo {author} {\bibfnamefont
  {M.}~\bibnamefont {Senne}}, \bibinfo {author} {\bibfnamefont
  {M.}~\bibnamefont {Held}}, \bibinfo {author} {\bibfnamefont {J.~D.}\
  \bibnamefont {Chodera}}, \bibinfo {author} {\bibfnamefont {C.}~\bibnamefont
  {Sch{\"{u}}tte}}, \ and\ \bibinfo {author} {\bibfnamefont {F.}~\bibnamefont
  {No{\'{e}}}},\ }\href@noop {} {\bibfield  {journal} {\bibinfo  {journal} {J.
  Chem. Phys.}\ }\textbf {\bibinfo {volume} {134}},\ \bibinfo {pages} {174105}
  (\bibinfo {year} {2011})}\BibitemShut {NoStop}%
\end{thebibliography}

%

\end{document}